\documentclass[pdflatex,sn-basic,iicol]{sn-jnl}% Math and Physical Sciences Numbered Reference Style
\usepackage{graphicx}%
\usepackage{multirow}%
\usepackage{amsmath,amssymb,amsfonts}%
\DeclareMathAlphabet{\mathbbold}{U}{bbold}{m}{n}
\usepackage{amsthm}%
\usepackage{mathrsfs}%
\usepackage[title]{appendix}%
\usepackage[table]{xcolor}%
\usepackage{textcomp}%
\usepackage{pifont}%
\usepackage{manyfoot}%
\usepackage{booktabs}%
\usepackage{algorithm}%
\usepackage{algorithmicx}%
\usepackage{algpseudocode}%
\usepackage{listings}%
\usepackage{graphicx}
\usepackage{stmaryrd}
\usepackage{siunitx}
\usepackage{etoolbox}

\makeatletter
\renewcommand{\addressfont}{\reset@font\fontsize{9bp}{11bp}\selectfont\titraggedcenter}
\patchcmd{\@maketitle}{\vskip7pt\addressfont}{\vskip5pt\addressfont}{}{}
\patchcmd{\@maketitle}{\removelastskip\vskip24pt}{\removelastskip\vskip12pt}{}{}
\patchcmd{\@maketitle}{\removelastskip\vskip24pt}{\removelastskip\vskip14pt}{}{}
\makeatother

\theoremstyle{thmstyleone}%
\theoremstyle{thmstyletwo}%

\theoremstyle{thmstylethree}%

\begin{document}

\title[Video Individual Counting and Tracking from Moving Drones: A Benchmark and Methods]{Video Individual Counting and Tracking from Moving Drones: A Benchmark and Methods}

\author[1]{\fnm{Yaowu} \sur{Fan}}

\author[2]{\fnm{Jia} \sur{Wan}}

\author[3]{\fnm{Tao} \sur{Han}}

\author*[1]{\fnm{Andy J.} \sur{Ma}}\email{majh8@mail.sysu.edu.cn}

\author[4]{\fnm{Wanli} \sur{Ouyang}}

\author[5]{\fnm{Antoni B.} \sur{Chan}}

\affil[1]{\orgdiv{School of Computer Science and Engineering},
	\orgname{Sun Yat-sen University},
	\orgaddress{\city{Guangzhou}, \postcode{510275}, \country{China}}}

\affil[2]{\orgdiv{School of Computer Science and Technology},
	\orgname{Harbin Institute of Technology (Shenzhen)},
	\orgaddress{\city{Shenzhen}, \postcode{518066}, \country{China}}}

\affil[3]{\orgdiv{School of Computer Science and Engineering},
	\orgname{Hong Kong University of Science and Technology},
	\orgaddress{\city{Hong Kong}, \country{China}}}

\affil[4]{\orgdiv{Department of Information Engineering},
	\orgname{Chinese University of Hong Kong},
	\orgaddress{\city{Hong Kong}, \country{China}}}

\affil[5]{\orgdiv{Department of Computer Science},
	\orgname{City University of Hong Kong},
	\orgaddress{\city{Hong Kong}, \country{China}}}


\abstract{Counting and tracking dense crowds in large-scale scenes is valuable yet challenging,
	while existing methods and datasets are largely limited to fixed cameras with small scene coverage.
	We introduce MovingDroneCrowd++, a large-scale video-level dataset dedicated to dense crowd counting and tracking from moving drones, captured under diverse flight altitudes, camera angles, and illumination conditions.
	Existing methods, however, still fail to achieve satisfactory Video Individual Counting (VIC) or Multi-Object Tracking (MOT) performance under these challenging aerial conditions.
	% To this end, we propose GD\textsuperscript{3}A (\underline{G}lobal \underline{D}ensity map \underline{D}ecomposition via Group-wise \underline{D}ensity  \underline{A}ssignment), a video individual counting method that first establishes pixel-level correspondences between pedestrian descriptors across frames via optimal transport with an frame-pair-conditioned dustbin score.
	% Then, group-wise density assignment is performed to guide the decomposition of global density map into shared, inflow, and outflow density maps. 
	% We further introduce a pedestrian tracking method, DVTrack (\underline{D}escriptor-\underline{V}oting Track), which converts descriptor-level matching into instance-level association through descriptor voting. 
	% Our methods rely on the association and voting results of group-wise multiple descriptors for each pedestrian rather than a single vector. Since intra-group matching errors do not affect the final counting and tracking results, our methods are more robust in dense crowds and challenging aerial conditions.
	To this end, we propose GD\textsuperscript{3}A (\underline{G}lobal \underline{D}ensity Map \underline{D}ecomposition via Group-wise \underline{D}ensity \underline{A}ssignment) for VIC and GIA-Track (\underline{G}roup-wise \underline{I}dentity \underline{A}ssociation Tracker) for MOT. Both methods are unified under a framework that aggregates cross-frame pixel-level matches among multiple descriptors per pedestrian through group-wise density assignment (GDA) and group-wise identity association (GIA), respectively. This group-wise aggregation tolerates intra-group spatial mismatches and limits the propagation of inter-group errors. To establish reliable pixel-level descriptor correspondences across frames, we design a frame-pair-conditioned dustbin score inferred from the pedestrian descriptors of each frame pair, enabling Optimal Transport to better distinguish identity differences. Based on these matches, GD\textsuperscript{3}A decomposes global density maps into shared, inflow, and outflow components, while GIA-Track establishes robust pedestrian trajectories.
	Experiments show that our methods achieve substantial gains in both VIC and MOT on moving-drone videos with dense crowds and complex motions, reducing counting error by 47.4\% and improving tracking accuracy by 64.6\%.
	Code, dataset, and pretrained models are available \href{https://github.com/fyw1999/MovingDroneCrowd}{here}.}




\maketitle

\section{Introduction}\label{sec1}

In recent years, with the rapid advancement of artificial intelligence, the low-altitude economy has experienced explosive growth as an emerging industry~\citep{low-altitude_economy}. UAVs (commonly referred to as drones) play a central role due to their mobility and flexibility. 
By integrating drones with crowd analysis algorithms, such as counting or tracking~\citep{LTCO, csrnet, P2P, Wang2021Pixel, Zhang2021FairMOT, Dendorfer2021MOTChallenge}, it becomes possible to perform flexible monitoring and density estimation of pedestrians in large-scale scenes, which effectively prevents crowd congestion and stampede-related accidents, and is of great significance
for public safety~\citep{public_safety}. 

\begin{figure*}[!t]
	\centering
	\includegraphics[width=\textwidth]{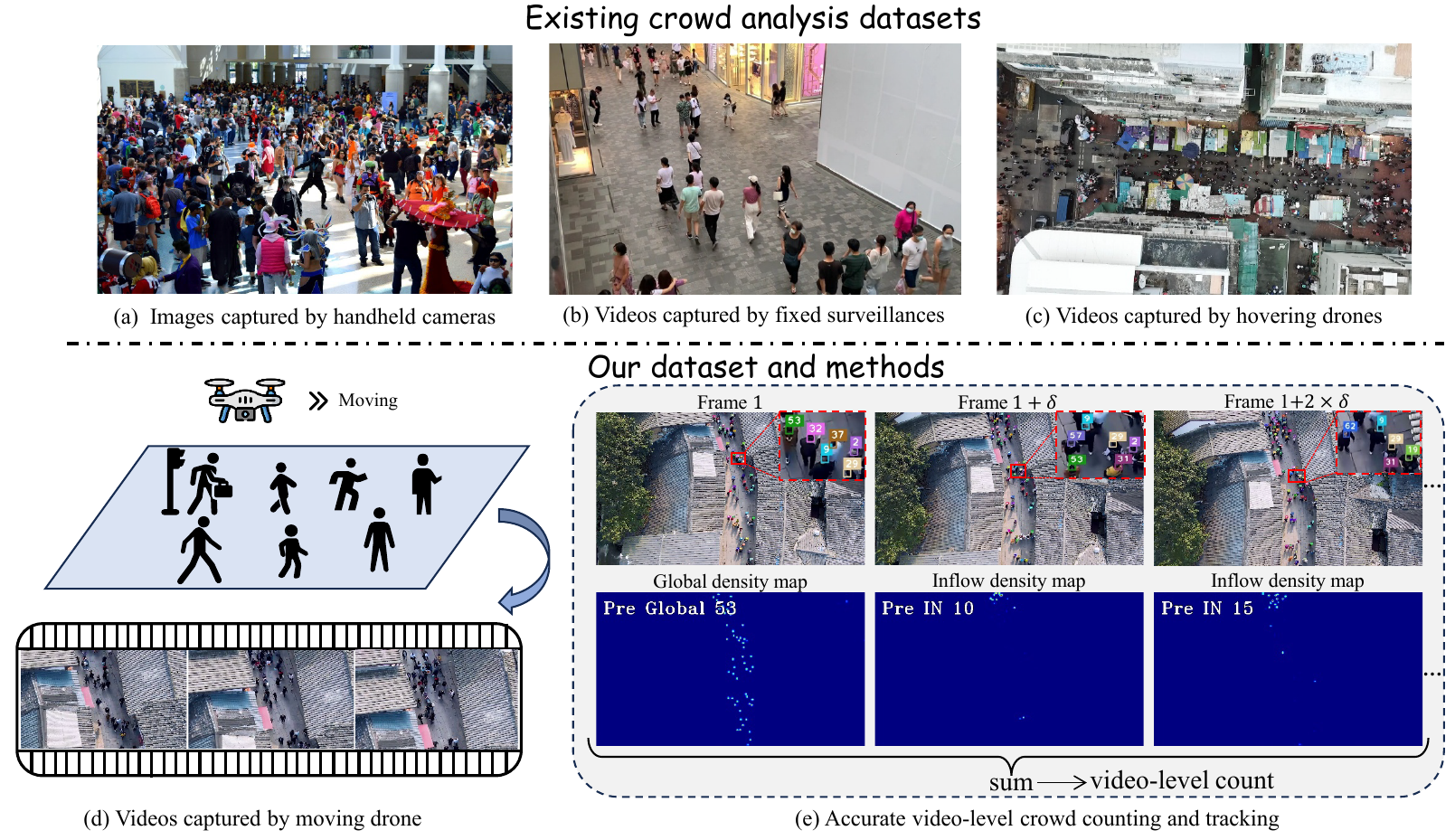}
	\caption{Comparison between existing crowd analysis datasets and ours. Existing research has predominantly focused on (a) free-viewpoint images captured by handheld cameras, (b) videos captured by fixed surveillance, or (c) hovering drones. Due to the constraints of these data acquisition setups, prior methods cannot perform video-level crowd counting and tracking in large-scale, crowded environments. Our method utilizes moving drones to capture videos covering large-scale scenes (d) and achieves accurate and interpretable video-level crowd counting and tracking (e).}
	\label{fig:motivation}
\end{figure*}

However, existing crowd analysis algorithms and datasets mainly focus on static images captured by handheld cameras~\citep{gao2020nwpu, ucf-qnrf, Zhang_2016_CVPR_MCNN, UCF-CC-50, JHU-CROWD++}, or on videos recorded by fixed surveillance cameras~\citep{VSCrowd, DanceTrack, Mot20, SportsMOT, WuhanMetroCrowd} and hovering drones~\citep{DroneCrowd, VisDrone-CC2021} (see Fig. \ref{fig:motivation} (a) $\sim$ (c)). Due to the limited mobility of the capturing devices, these data can only cover crowds within fixed areas, making them unsuitable for counting or tracking dense crowds in large-scale scenes. In contrast, videos captured by moving drones enable both video individual counting (VIC) and tracking over large-scale scenes, where VIC aims to estimate the number of unique pedestrians appearing throughout an entire video.
Although several related datasets, such as UAVVIC~\citep{uavvic} and VisDrone~\citep{VisDroneDatasets}, have been proposed, they suffer from several significant limitations. 
Most videos in UAVVIC are still captured by hovering drones with limited fields of view, and neither dataset focuses on dense crowds, with dense crowds in VisDrone even labeled as ignore regions. Moreover, their videos are mainly recorded in suburban areas with sparse crowds and limited diversity in flight altitude, viewing angle, and illumination, leading to a large domain gap from real-world dense crowd scenarios.
As a result, no existing dataset simultaneously satisfies all these requirements: \textbf{dense crowds, diverse and complex environments, highly mobile drone-based acquisition, and large-scale scene coverage}.

Beyond dataset limitations, accurately counting or tracking dense crowds in videos captured by highly dynamic drones remains challenging. Existing multi-object tracking (MOT) methods~\citep{ByteTrack, DanceTrack, TrackFormer, MOTIP, SparseTrack, DiffusionTrack} are generally effective only in simple scenarios with few and relatively large targets, but their performance degrades severely in dense crowds and under complex motion. 
VIC methods~\citep{DRNet, PGDTR, uavvic} are proposed to decompose video-level counting into estimating the number of pedestrians in the initial frame and the inflow pedestrian count for each subsequent frame.
% However, current methods heavily rely on accurate localization and strict one-to-one association, which are extremely difficult in dense crowds. As a result, localization and association errors accumulate and significantly degrade counting accuracy. Although density map-based VIC methods~\citep{MDC, FMDC} partially alleviate this issue, directly regressing the inflow, outflow, or shared density map remains challenging and lacks interpretability. Moreover, methods that compute cross-attention between feature maps of adjacent frames to estimate inflow density map~\citep{MDC, FCCF} incur high computational costs, making them unsuitable for efficient deployment in real-world applications.
However, current methods heavily rely on accurate localization followed by strict one-to-one association, a paradigm that is particularly fragile in dense crowds because a single incorrect cross-frame assignment may incorrectly pair two pedestrians, exclude their correct partners, and consequently induce cascading mismatches among the remaining pedestrians. As a result, localization and association errors accumulate and significantly degrade counting accuracy. Although density map-based VIC methods~\citep{MDC, FMDC} partially alleviate this issue, directly regressing the inflow, outflow, or shared density map remains challenging and lacks interpretability. Moreover, methods that compute cross-attention between feature maps of adjacent frames to estimate the inflow density map~\citep{MDC, FCCF} incur high computational costs, making them unsuitable for efficient deployment in real-world applications.

\begin{figure*}[!t]
	\centering
	\includegraphics[width=\textwidth]{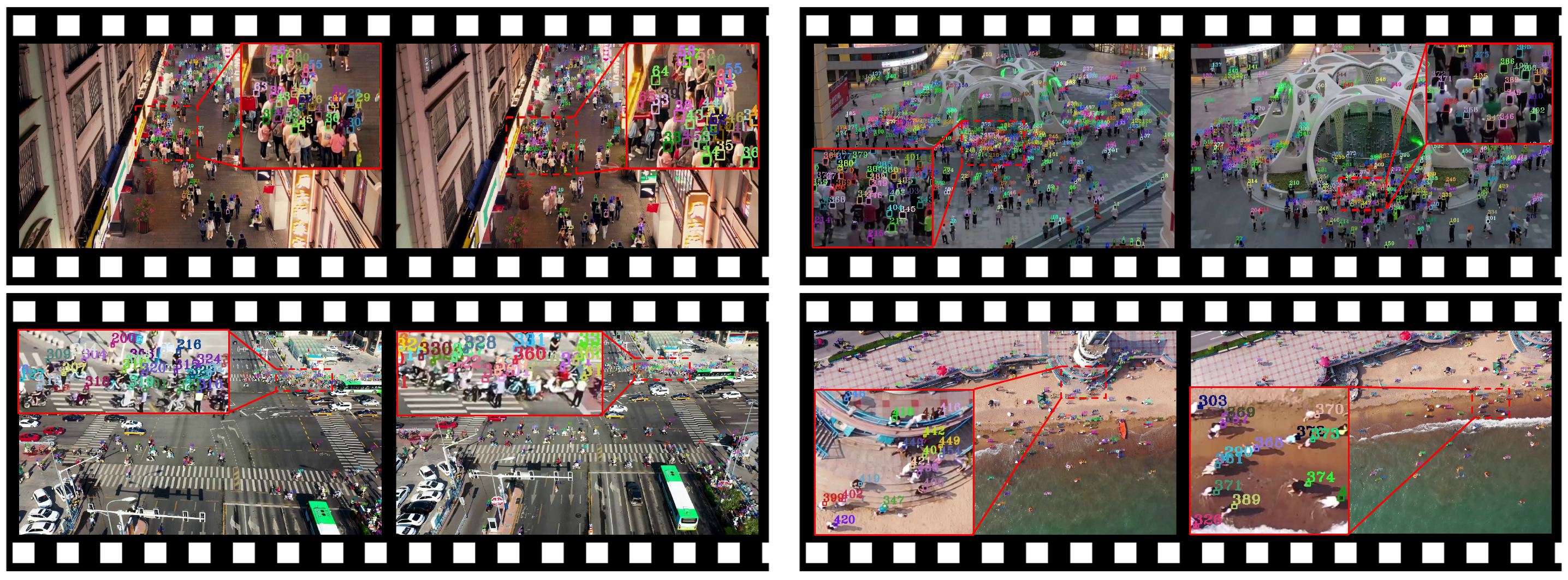}
	\caption{Exemplars from the MovingDroneCrowd++ dataset. Due to space constraints, only two frames are displayed for each video clip. Each frame is annotated with a bounding box and an identity ID for every pedestrian head. These examples illustrate that the dataset is captured by moving drones in dense crowd environments and exhibits significant diversity in terms of shooting angles, flight altitudes, and illumination conditions.}
	\label{fig:dataset_examples}
\end{figure*}

In this paper, we study a practical yet underexplored problem: ``\textbf{\textit{How to achieve accurate and efficient crowd counting and tracking in large-scale scenes with dense crowds?}}'' To address this and overcome the aforementioned limitations, we introduce a large-scale dataset for video individual counting and tracking from moving drone perspectives in large-scale scenes with dense crowds, comprising 638,718 head bounding boxes and 27,866 pedestrian trajectories. Unlike previous datasets that are constrained by limited fields of view or simple acquisition conditions, MovingDroneCrowd++ exhibits three key characteristics: high dynamics, dense crowds, and diverse and complex acquisition conditions. It encompasses various lighting conditions, shooting angles, and flight
altitudes, bridging the domain gap and capturing the complexity of real-world environments to support model training and evaluation (see Fig. \ref{fig:motivation} (d) and Fig. \ref{fig:dataset_examples}).

Due to the diversity and complexity of our dataset, existing counting and tracking methods struggle to handle its challenging scenarios effectively. To address these challenges, we propose \textbf{GD\textsuperscript{3}A} (\underline{\textbf{G}}lobal \underline{\textbf{D}}ensity Map \underline{\textbf{D}}ecomposition via Group-wise \underline{\textbf{D}}ensity \underline{\textbf{A}}ssignment) for VIC and \textbf{GIA-Track} (\underline{G}roup-wise \underline{I}dentity \underline{A}ssociation Tracker) for MOT. Rather than treating the two tasks independently, both VIC and MOT are formulated within a unified framework built on cross-frame pixel-level descriptor correspondences. The framework first uses predicted global density maps to filter out background descriptors and retain multiple descriptors for each pedestrian head, thereby substantially reducing the computational cost of pixel-level matching. It then establishes cross-frame correspondences among the retained descriptors via optimal transport~\citep{peyre2019computational} with a frame-pair-conditioned dustbin score inferred from the pedestrian descriptors of each input frame pair to improve cross-frame identity discrimination. The resulting OT matching matrix provides descriptor-level matching evidence rather than final pedestrian-level assignments.

Based on the same descriptor-level matching evidence, GD\textsuperscript{3}A performs group-wise density assignment (GDA) to decompose global density maps into shared, inflow, and outflow density maps, enabling efficient, accurate, and interpretable VIC, whereas GIA-Track performs group-wise identity association (GIA) to establish pedestrian identities and trajectories without additional training. Although GDA and GIA produce different task outputs, both aggregate multiple descriptor correspondences for each pedestrian. This group-wise aggregation tolerates intra-group spatial mismatches, such as correspondences between different locations of the same pedestrian head, and limits the propagation of isolated inter-group descriptor errors. Specifically, a single cross-identity descriptor mismatch contributes only partial erroneous evidence and does not by itself determine an entire pedestrian association or constrain the assignments of other pedestrians. Through pixel-level descriptor matching followed by group-wise evidence aggregation, our methods achieve accurate and robust VIC and MOT in dense crowds under highly dynamic drone motion.

The contributions of this paper are summarized as follows:

\begin{list}{\textbullet}{%
	\setlength{\leftmargin}{1.4em}%
	\setlength{\labelwidth}{0.6em}%
	\setlength{\labelsep}{0.4em}%
	\setlength{\itemindent}{0pt}%
	\setlength{\listparindent}{0pt}%
	\setlength{\rightmargin}{0pt}%
	\setlength{\topsep}{4pt}%
	\setlength{\partopsep}{0pt}%
	\setlength{\parsep}{0pt}%
	\setlength{\itemsep}{3pt}%
}
	\item We introduce MovingDroneCrowd++, a diverse and challenging dataset specifically designed for VIC and MOT in large-scale crowded scenes captured by moving drones under varied heights, angles, and lighting.
	
	% \item We propose GD\textsuperscript{3}A, which formulates inflow density estimation as a global density map decomposition problem through interpretable pixel-level pedestrian descriptor association. It excludes dominant background pixels to reduce computational cost and mitigate error accumulation caused by mismatches.
	
	% \item We propose GD\textsuperscript{3}A, which enables efficient, accurate, and interpretable VIC in challenging moving-drone scenarios with dense crowds by decomposing global density maps through robust group-wise density assignment. This assignment is based on pixel-level descriptor matches established via optimal transport with a frame-pair-conditioned dustbin score.
	% \item \textcolor{red}{We formulate VIC and MOT within a unified framework built on pixel-level descriptor correspondences. Within this framework, we propose GD\textsuperscript{3}A, which enables efficient, accurate, and interpretable VIC in challenging moving-drone scenarios with dense crowds by aggregating descriptor-level matching evidence through GDA to decompose global density maps. The descriptor correspondences are established via optimal transport with a frame-pair-conditioned dustbin score.}
	\item We formulate VIC and MOT within a unified framework built on cross-frame pixel-level descriptor correspondences. To establish reliable correspondences, we introduce a frame-pair-conditioned dustbin score inferred from the pedestrian descriptors of each input frame pair, enabling optimal transport to improve cross-frame identity discrimination.
	
	% \item We further propose GIA-Track, which converts pixel-level matches into instance-level associations by aggregating matches between multiple pedestrian descriptors across adjacent frames through group-wise identity association, delivering superior tracking performance in dense crowd scenes under highly dynamic drone motion.
	% \item \textcolor{red}{Using the same pixel-level descriptor correspondences, we further propose GIA-Track, which aggregates descriptor-level matching evidence through GIA to obtain pedestrian-level identity associations, delivering superior tracking performance in dense crowd scenes under highly dynamic drone motion. Together, GDA and GIA improve tolerance to intra-group spatial mismatches and limit the propagation of isolated inter-group descriptor errors.}
	% \item \textcolor{red}{We propose GD\textsuperscript{3}A and GIA-Track, which respectively decompose global density maps through GDA and establish pedestrian trajectories through GIA. Aggregating multiple correspondences per pedestrian makes both methods tolerant to intra-group spatial mismatches and limits the propagation of isolated inter-group descriptor errors.}
	\item We propose GD\textsuperscript{3}A and GIA-Track, which enable accurate and robust VIC and MOT in dense crowds under highly dynamic drone motion. Based on descriptor-level matches, GD\textsuperscript{3}A decomposes global density maps through the proposed GDA, whereas GIA-Track establishes pedestrian trajectories through the proposed  GIA. Both mechanisms aggregate multiple descriptor correspondences per pedestrian, thereby tolerating intra-group spatial mismatches and limiting the propagation of inter-group errors.
	
	\item Extensive experiments demonstrate the superiority of our methods. GD\textsuperscript{3}A and GIA-Track substantially outperform previous methods in moving drone scenarios with dense crowds and complex motions, reducing the counting error by \textbf{47.4\%} and improving tracking performance by \textbf{64.6\%}.
\end{list}

This work extends our preliminary research~\citep{MDC} in four key aspects: 1) First, we nearly double the size of the original dataset, increasing the number of head bounding-box annotations from 325,542 to 638,718 by incorporating large-scale, densely crowded, and low-light scenes. This expansion creates a more diverse and challenging benchmark that accurately represents the complexity of
real-world environments with dense crowds.
2) Second, we extend the task scope of the conference version from video individual counting (VIC) to both VIC and multi-object tracking (MOT) and develop a unified framework for the two tasks built on cross-frame pixel-level descriptor correspondences. To establish reliable correspondences, we introduce a frame-pair-conditioned dustbin score inferred from the pedestrian descriptors of each input frame pair, enabling optimal transport to improve cross-frame identity discrimination. 3) Third, within this framework, we propose GD\textsuperscript{3}A for VIC and GIA-Track for MOT. GD\textsuperscript{3}A decomposes global density maps into shared, inflow, and outflow components through group-wise density assignment, whereas GIA-Track establishes pedestrian trajectories through group-wise identity association. Compared with our conference method SDNet, GD\textsuperscript{3}A achieves higher counting accuracy and computational efficiency together with improved interpretability, while GIA-Track inherits these advantages from the shared framework and extends them to dense-crowd tracking without additional training.
4) Finally, we conducted more comprehensive experiments and qualitative visualizations to validate the effectiveness and interpretability of the proposed methods. The results demonstrate that our methods significantly outperform prior works in both VIC and MOT (reducing the counting error by 47.4\% and improving tracking performance by 64.6\%).

The remainder of this paper is organized as follows. 
Section~2 reviews related work. 
Section~3 introduces the MovingDroneCrowd++ dataset. 
Section~4 presents the proposed GD\textsuperscript{3}A and GIA-Track methods. 
Section~5 reports experimental results and analyses. 
Section~6 concludes this paper.

\section{Related Work}\label{sec2}
\subsection{Image-level Crowd Counting}
Image-level crowd counting aims to estimate the number of people in a given static image~\citep{csrnet, Zhang_2016_CVPR_MCNN, Liu_2019_CVPR_Context_Aware, Reverse_Perspective, wang2020DMCount, STEERER, CrowdDiff}. As a fundamental task in computer vision, it plays a crucial role in many real-world applications. This field has undergone substantial evolution and development over the past several years. In the early stages, detection-based methods~\citep{KRR} were sufficient for crowd counting in sparse scenarios. To better tackle dense crowd scenarios, regression-based approaches~\citep{PPCM, LTCO} were subsequently introduced. With the prevalence of data-driven deep learning, datasets featuring extremely dense crowds have been introduced~\citep{gao2020nwpu, JHU-CROWD++, UCF-CC-50, ucf-qnrf}. Density map estimation-based methods~\citep{ Zhang_2016_CVPR_MCNN, Liu_2019_CVPR_Context_Aware}, which exhibit superior performance in such highly crowded scenes, have consequently become the dominant approach. As the complexity of the data increased, the field began to face several new challenges, including perspective effects~\citep{Reverse_Perspective, Perspective-Guided}, head scale differences~\citep{STEERER, Redesigning_Multi-Scale}, and domain gaps~\citep{HQITDR, Domain-General_Du_Deng_Shi_2023, Striking_a_Balance}. Moreover, researchers have proposed new loss functions~\citep{wan2021generalized, wang2020DMCount, Bayesian_Poisson}, network architectures~\citep{P2P, E2ETCL}, and supervision strategies~\citep{LCSD, wan2020modeling, 9189836, ADMGCC}. Multi-view crowd counting~\citep{CountFormer, SynMVCrowd} extends crowd counting to larger-scale scenes to some extent, but it is still limited to fixed locations. Recently, Embodied Crowd Counting (ECC)~\citep{ECC} has been proposed for actively counting pedestrians in large-scale scenes. However, existing ECC methods rely on synthetic data and suffer from a significant domain gap with complex real-world environments, which limits their application. Although these studies have significantly advanced image-level crowd counting, the limited field of view and inflexibility of static images greatly constrain their applicability in the real-world, particularly in large-scale scenes with dense crowds.
\subsection{Video-level Crowd Counting and Multi-Object Tracking}
Video-level crowd counting~\citep{MDC, uavvic, DRNet, PGDTR, FMDC, FCCF}, defined as Video Individual Counting (VIC) in~\citet{DRNet}, aims to estimate the number of unique pedestrians across an entire video. 
% As an emerging subfield of crowd counting, it has attracted considerable attention in recent years. 
% \citet{DRNet} first modeled this task as predicting the number of pedestrians in the first frame and the inflow count in each subsequent frame.~\citep{uavvic} proposed a weakly supervised approach that guides the learning process using predicted similarity. Other works have introduced density map-based methods~\citep{MDC, FMDC, FCCF}, with~\citep{FMDC} directly predicting inflow and outflow density maps, while~\citep{MDC} first predicts shared density maps and then derives the inflow and outflow density maps by subtracting the shared density maps from the global density maps. Localization-based methods rely on precise localization and strict one-to-one association, whereas density map-based methods either suffer from poor interpretability or incur high computational costs due to cross-frame attention. In contrast, multi-object tracking~\citep{DTLD_MOT, DeconfuseTrack, UTM} is a classical task, whose common paradigm is tracking-by-detection~\citep{MOTIP, BoT_SORT, DiffMOT, OC-SORT}, where detection results are associated with historical trajectories using Kalman filtering~\citep{Kalman_Filter}. Recently, some methods~\citep{MOTIP, TrackFormer, MOTR} have adopted Transformer and employ track queries to track targets. However, these methods cannot effectively handle dense crowds and fast drone motion. In contrast, our methods employ GDA and GIA to achieve robust group-wise reasoning with strong tolerance to errors in dense scenes and complex conditions.
\citet{DRNet} first modeled this task as predicting the number of pedestrians in the first frame and the inflow count in each subsequent frame.~\citep{uavvic} proposed a weakly supervised approach that guides the learning process using predicted similarity. Other works have introduced density map-based methods~\citep{MDC, FMDC, FCCF}, with~\citep{FMDC} directly predicting inflow and outflow density maps, while~\citep{MDC} first predicts shared density maps and then derives the inflow and outflow density maps by subtracting the shared density maps from the global density maps. Localization-based methods rely on precise localization and strict one-to-one association, whereas density map-based methods either suffer from poor interpretability or incur high computational costs due to cross-frame attention. In contrast, Multi-Object Tracking (MOT)~\citep{DTLD_MOT, DeconfuseTrack, UTM} is a classical task, whose common paradigm is tracking-by-detection~\citep{MOTIP, BoT_SORT, DiffMOT, OC-SORT}, where detection results are associated with historical trajectories using Kalman filtering~\citep{Kalman_Filter}. Recently, some methods~\citep{MOTIP, TrackFormer, MOTR} have adopted Transformer and employ track queries to track targets. However, these methods cannot effectively handle dense crowds and fast drone motion.
% \textcolor{red}{In contrast, we formulate VIC and MOT within a unified framework built on pixel-level descriptor correspondences. Rather than directly committing to one-to-one pedestrian assignments, GDA and GIA aggregate descriptor-level matching evidence to obtain task-specific density and identity assignments, thereby reducing the propagation of matching errors.}
In contrast, we formulate VIC and MOT within a unified framework built on pixel-level descriptor correspondences. By aggregating multiple descriptor correspondences for each pedestrian instead of directly committing to strict one-to-one pedestrian assignments, the proposed framework tolerates intra-group spatial mismatches and limits the propagation of inter-group descriptor errors.

% Declare this full-width float while page 5 is being built so that it can
% appear at the top of page 6.

% Define Table 1 here, but enqueue it later while page 6 is being built.
\newcommand{\placedatasetcomparisontable}{%
\begin{table*}[!t]
	\centering
	\caption{Comparison with related datasets. MovingDroneCrowd++ is a large-scale dataset dedicated to
		video individual counting and tracking in moving drones scenarios with dense crowds. It exhibits the most significant diversity in shooting angles, altitudes, and illumination conditions. These factors, combined with its high dynamic characteristics, make it highly challenging.}
	\label{tab:dataset_comparison}
	\resizebox{\linewidth}{!}{
		\begin{tabular}{ccccccccccccc}
			\hline
			Dataset 
			& Perspective  & Resolution & Moving & Images & Dynamic Frames 
			& Scenes & Boxes & Tracks & Light & Height 
			& Angle & IDs \\ \hline
			CroHD~\citep{Headhunter-T}
			& Surveillance & 1080P & \ding{55} &11,464 & 0
			& 5 & 1,188,496 & 2,752 & day\&night & Fixed 
			& Fixed &  \checkmark \\
			VSCrowd~\citep{VSCrowd}
			& Surveillance & 4K-360P & \ding{55}  & 62,938 & 0
			& 153 & 2,011,551 & 43,179 & day\&night &  Fixed 
			& Fixed & \checkmark \\ 
			WuhanMetroCrowd~\citep{WuhanMetroCrowd}
			& Surveillance & 1080P-720P & \ding{55}  & 11,925 & 0
			& 15  & 223,662 & -- & -- &  Fixed 
			& Fixed & \ding{55} \\ \hline
			DroneCrowd~\citep{DroneCrowd}
			& Drone & 1080P & \ding{55} &33,600 & 0
			& 25 & 4,864,280 & 20,800 & day\&night & Fixed
			& Fixed & \checkmark \\ 
			VisDrone~\citep{VisDroneDatasets}
			& Drone & 4K-360P & \checkmark\kern-1.1ex\raisebox{.7ex}{\rotatebox[origin=c]{125}{--}} & 33,682 & 64\%
			& 57 & 519,196 & 3,976 & day\&night & $\sim$ 10m 
			& $\sim$ 45--90\textdegree{} & \checkmark \\ 
			UAVVIC~\citep{uavvic}
			& Drone & 4K-1080P & \checkmark\kern-1.1ex\raisebox{.7ex}{\rotatebox[origin=c]{125}{--}} & 5,396 & 51\% 
			& 24 & 398,158 & -- & day & $\sim$ 20m 
			& $\sim$ 90\textdegree{} & \ding{55} \\
			MovingDroneCrowd~\citep{MDC}
			& Drone & 4K-720P & \checkmark & 4,940 & 100\% 
			& 26 & 325,542 & 16,154 & day\&night &  $\sim$ 3-20m
			&  $\sim$ 45--90\textdegree{} & \checkmark \\ 
			MovingDroneCrowd++ 
			& Drone & 4K-720P & \cellcolor{yellow}\checkmark & \cellcolor{yellow}7,197 & \cellcolor{yellow}100\%
			& \cellcolor{yellow}44 & \cellcolor{yellow}638,718 & \cellcolor{yellow} 27,866 & \cellcolor{yellow}day\&night & \cellcolor{yellow}$\sim$ 3-20m
			& \cellcolor{yellow}$\sim$ 45--90\textdegree{} & \cellcolor{yellow}\checkmark \\ \hline
		\end{tabular}
	}
\end{table*}
}

\subsection{Drone-based Crowd Counting and Tracking}
To overcome the limitations of ground-based cameras, such as handheld devices and surveillance cameras, drones have also been employed in crowd counting and tracking due to their high flexibility~\citep{VisDrone-CC2021, RGB-T_CCD, DroneCrowd, DBJDLT, WANG2026112778, drones9120833, Drone_Person_Tracking, DenseTrack, MFAFLW}. However, these methods remain confined to image-level crowd counting~\citep{RGB-T_CCD, drones9120833} or tracking on fixed-view videos captured by hovering drones~\citep{DenseTrack, MFAFLW}. UAVVIC~\citep{uavvic} contains only a small number of videos captured by moving drones, with most clips recorded in suburban areas featuring sparse crowds and limited variations in camera angle, altitude, and illumination. Consequently, it shows a substantial domain gap from real-world dense crowd scenarios. VisDrone~\citep{VisDroneDatasets} is a large-scale MOT dataset, but dense crowds are deliberately annotated as ignore regions. Our conference paper~\citep{MDC} introduces MovingDroneCrowd, a video dataset captured by moving drones in complex environments with dense crowds, featuring diverse variations in shooting angles, flight altitudes, and illumination conditions. However, it remains limited in scale and lacks large-scale, long-duration videos, making it insufficient for thoroughly evaluating the performance of different algorithms. 
MovingDroneCrowd++ introduces longer dynamic drone videos and nearly doubles the annotation scale of MovingDroneCrowd, making it the most diverse and challenging dynamic drone video dataset for dense crowd scenarios to date.

\section{MovingDroneCrowd++}
\subsection{Data Collection and Processing}
\subsubsection{Collection} Due to the strict regulations on drone operations in crowded environments, acquiring dynamic drone video data in these areas presents a significant challenge, particularly in the highly congested scenarios targeted in this work, such as commercial districts, pedestrian streets, and tourist attractions. To this end, we curated publicly available drone videos depicting crowded outdoor public spaces. Candidate raw videos were identified through publicly accessible video platforms and search engines using crowd-related aerial-view keywords, such as ``aerial crowd,'' ``aerial pedestrian street,'' and ``aerial tourist attractions.'' Subsequently, we selected and downloaded videos that adhere to the following criteria:
1) The video clips must be captured by moving drones over crowded scenes.
2) The video content must be pedestrian-centric with distinguishable head features. This necessitates a moderate flight altitude and a sufficient depression angle to reduce occlusion. 

% Enqueue Table 1 late enough to avoid sharing page 6 with Fig. 2, but before
% page 7 is assembled, so it appears at the top of page 7.
\placedatasetcomparisontable

\subsubsection{Processing} For the downloaded videos, we first used \href{https://www.boilsoft.com/videosplitter}{\textcolor[rgb]{0.21,0.49,0.74}{Boilsoft Video Splitter}} to segment them into multiple continuous clips, with each clip covering a specific region as completely as possible. Videos that were already continuous and coherent were left unsegmented. Note that the clips extracted from the same original video are grouped into the same scene. This is because these clips are typically captured in the same or nearby areas under similar collection conditions, resulting in a consistent domain style. Subsequently, to eliminate redundancy and reduce annotation costs, we performed frame sampling on each clip. However, since the drone flight speeds varied across the original videos, we adjusted the sampling rate accordingly and downsampled the frame rate by factors of 3, 6, or 9, depending on the drone's motion speed in each clip. Finally, for clips with small depression angles, we cropped each frame around its center with a smaller resolution. This removes distant regions where pedestrians are difficult to distinguish, which simultaneously reduces annotation difficulty and uncertainty while increasing the amount of pedestrian inflow. In general, this cropping operation has little impact on the total number of people in the clip, since pedestrians that are cropped will re-enter the field of view as the drone moves forward.

\subsection{Dataset Annotation and Split}
\subsubsection{Instance-level Annotations} After completing the data collection and processing described above, the obtained video frames were assigned to 20 experienced annotators. During the annotation process, the annotators marked each pedestrian starting from the first frame in which the individual appeared. They labeled head bounding boxes that tightly enclose the head and assigned a unique identity across the entire clip, continuing the annotation until the pedestrian completely exited the view. If the pedestrian's head became occluded, annotation was temporarily suspended, and the same identity was reassigned once the pedestrian reappeared. A video clip is considered fully annotated once all pedestrians with distinct identities have been completely annotated from their first to their last visible frame. 

After the initial annotation of all video clips is completed, the annotations are reassigned to another group of annotators. For each trajectory, the inspection started from the first frame in which the pedestrian appeared and continued until the trajectory ended. Any errors identified during this process were recorded and corrected. \href{https://github.com/darkpgmr/DarkLabel}{\textcolor[rgb]{0.21,0.49,0.74}{Darklabel}} and \href{https://www.cvat.ai}{\textcolor[rgb]{0.21,0.49,0.74}{CVAT}} were used for the annotation process, while \href{https://github.com/sgumhold/TmoTA}{\textcolor[rgb]{0.21,0.49,0.74}{TmoTA}} was employed for verification. TmoTA provides a visualization of all pedestrian trajectories and can highlight the selected trajectory, which greatly facilitates efficient and accurate error inspection.

\subsubsection{Scene-level Annotations} In addition to the instance-level annotations, we also provide scene-level annotations for each video clip, including shooting time (daytime or nighttime), location, and difficulty level. The difficulty level is determined by the number of distinct pedestrians appearing in the entire clip, divided into four levels with intervals of 200 individuals. Moreover, compared to the conference version, the newly introduced video clips are additionally annotated with their durations. The scene-level annotations offer a principled foundation for both dataset splits and the evaluation of experimental results.

In total, we obtained \textbf{120} video clips from \textbf{44} distinct scenes, comprising \textbf{7,197} frames, \textbf{638,718} head bounding boxes, and \textbf{27,866} pedestrian trajectories. The complex and diverse acquisition conditions make MovingDroneCrowd++ the most diverse and challenging dataset to date specifically designed for video individual counting and tracking in dense crowded scenes captured by moving drones.

\subsubsection{Dataset Split} We split the dataset into training, validation, and test sets. Our dataset split has the following two important characteristics: \textbf{1) Scene-level Split}. This scene-level partition ensures that no video clips from the same or similar scenes appear across different subsets. This means that training and evaluation on our dataset are conducted in a cross-scene manner, which imposes a stronger requirement on the generalization capability of the algorithms. \textbf{2) Balanced Split}. With the scene-level annotations, we can perform a reasonable and balanced dataset split. This prevents undesirable biases, such as the training set contains most of the challenging scenes, while the test set mainly consists of simpler ones, which will distort evaluation results. Specifically, we split the dataset based on two main scene-level attributes: difficulty level and illumination. We first categorized all scenes according to these two key attributes and then randomly assigned the scenes within each category to the training, validation, and test sets following a predefined ratio.

\begin{figure}[!t]
	\centering
	\includegraphics[width=\columnwidth]{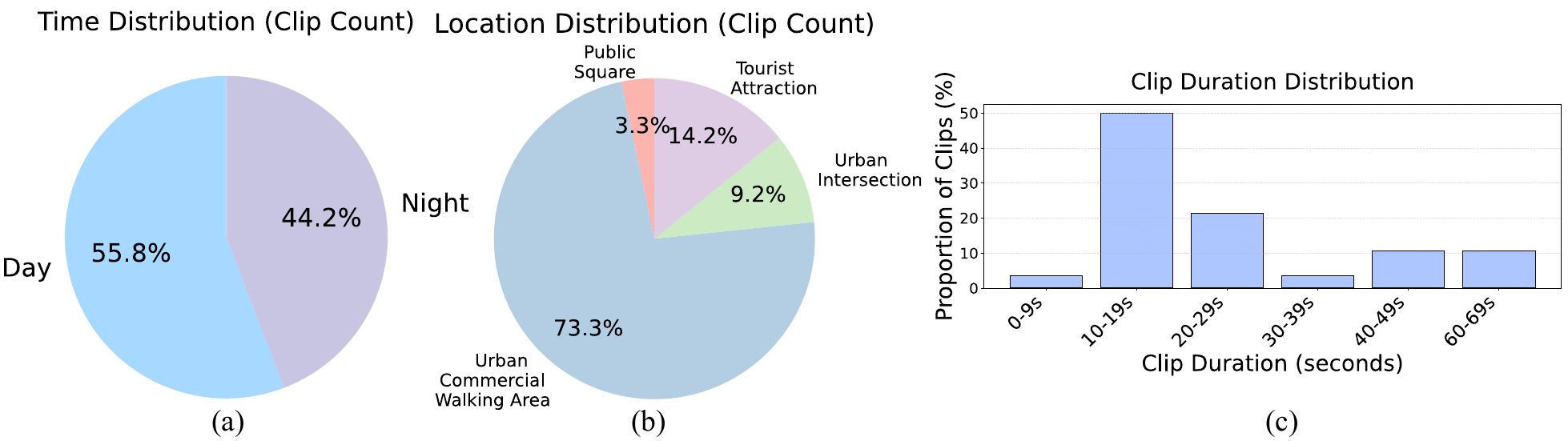}
	\caption{ Scene attributes statistics of the MovingDroneCrowd++ dataset.  (a) Proportion of illumination conditions. (b) Proportion of shooting locations. (c) Duration histogram of the newly added clips. These scene attributes statistics highlight the diversity and challenging nature of the proposed dataset.}
	\label{fig:scene_attributes_statistics}
\end{figure}

\subsection{Dataset Statistical Analysis and Comparison}
\subsubsection{Statistical Analysis} 
\begin{figure}[!t]
	\centering
	\includegraphics[width=\columnwidth]{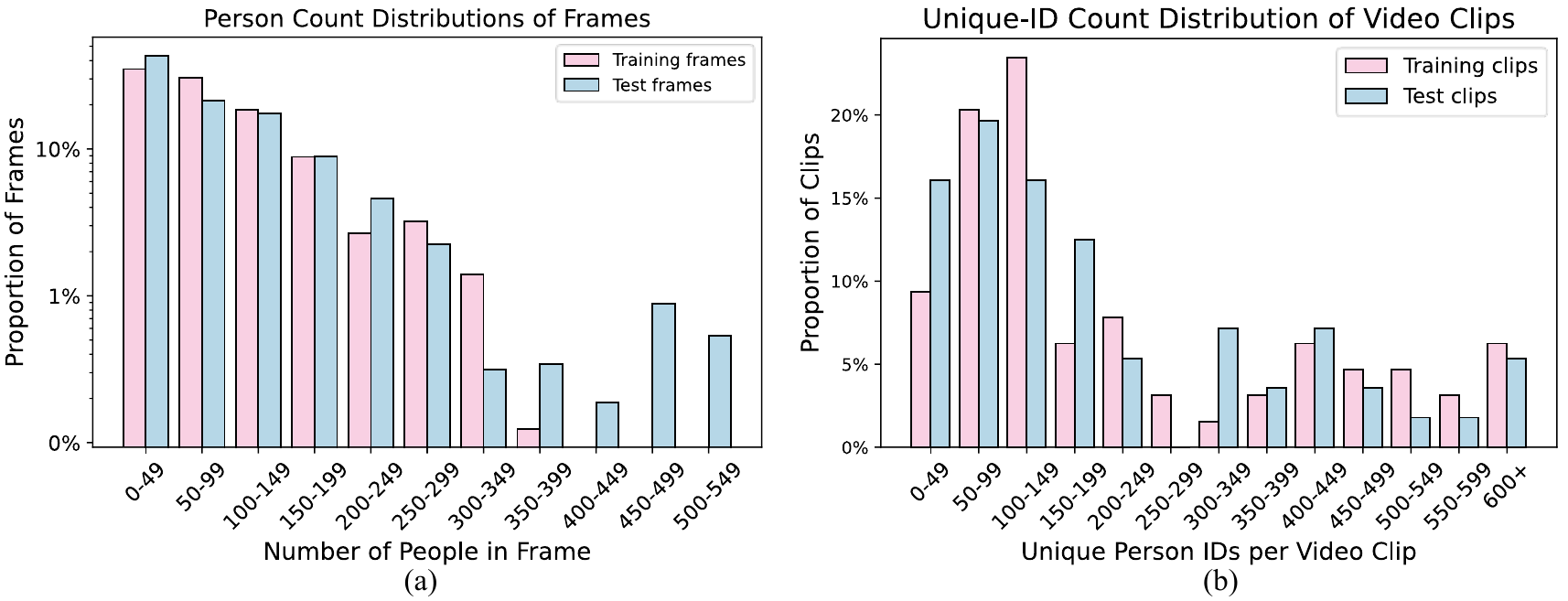}
	\caption{Crowd density statistics of the MovingDroneCrowd++ dataset. (a) Histogram of people per frame. (b) Histogram of distinct identities per clip. These density statistics demonstrate the balance of the dataset split.}
	\label{fig:crowd_density_statistics}
\end{figure}

Fig. \ref{fig:crowd_density_statistics} presents the count distribution of our dataset. Fig. \ref{fig:crowd_density_statistics}(a) and Fig. \ref{fig:crowd_density_statistics}(b) show the histograms of the number of pedestrians per frame and the number of distinct identities per video clip, respectively. As observed from these histograms, the distributions of the training and test sets are well balanced, enabling a fair and reliable evaluation of different algorithms. In addition, the histograms indicate that both the number of pedestrians per frame and the number of trajectories per clip are relatively dense. This demonstrates that our dataset effectively reflects the high-density pedestrian flows commonly observed in urban environments. Fig. \ref{fig:scene_attributes_statistics}(a) and Fig. \ref{fig:scene_attributes_statistics}(b) illustrate the distributions of shooting times and locations. Fig. \ref{fig:scene_attributes_statistics}(a) shows that the dataset contains a balanced number of videos captured during the daytime and at night. In particular, the inclusion of night-market scenes, which are classic examples of high-density pedestrian scenes under low-light conditions, further enhances the diversity and challenge of the dataset. Since pedestrian streets and commercial districts are difficult to distinguish clearly, we group them together as ``Urban Commercial Walking Area'' in Fig. \ref{fig:scene_attributes_statistics}(b). In addition, our dataset includes other typical high-density pedestrian areas, such as tourist attractions, intersections, and public squares.  Finally,  Fig. \ref{fig:scene_attributes_statistics}(c) presents the duration distribution of the newly added video clips. The distribution indicates that these new clips cover relatively large spatial areas, which further enriches the dataset.

\subsubsection{Comparison} Table \ref{tab:dataset_comparison} presents a comparison between our dataset and other related video datasets. Although our dataset is not the largest in scale, it surpasses fixed-camera datasets in terms of dynamic motion and difficulty, offering a much broader spatial coverage. Compared with the other drone-based video dataset such as UAVVIC, our dataset has clear advantages in dynamic motion, scale, and diversity of shooting conditions, including scene types, shooting angles, flight altitudes, and illumination. It provides a more faithful representation of complex and crowded scenes in challenging dynamic drone scenarios. Moreover, the annotation of pedestrian trajectories in our dataset enables the training and evaluation of more advanced and powerful algorithms.

\section{Methodology}

\begin{figure*}[!t]
	\centering
	\includegraphics[width=\textwidth]{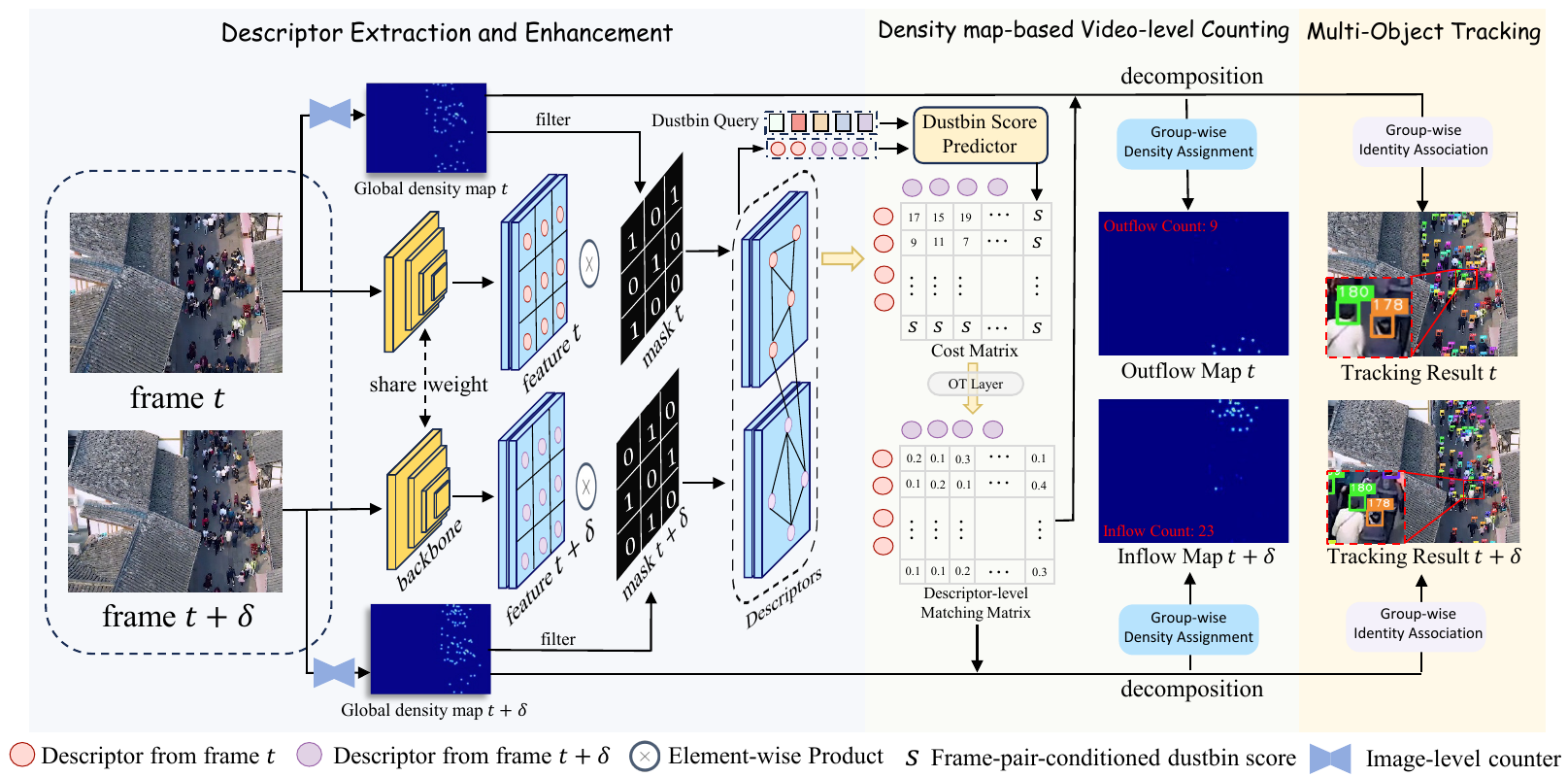}
	% \caption{The overall pipeline of the proposed GD\textsuperscript{3}A and DVTrack. Given two frames $F_t$ and $F_{t+\delta}$, a backbone extracts feature maps $\mathbf{F}_t$ and $\mathbf{F}_{t+\delta}$, which are filtered using global density maps $\hat{\mathbf{D}}^g_t$ and $\hat{\mathbf{D}}^g_{t+\delta} $ predicted by a pre-trained image-level estimator to retain visual descriptors for pedestrian heads. Subsequently, these descriptors are enhanced with positional coordinates and refined by an AGNN for contextual aggregation. Pixel-level matching between descriptors from two frames are established via Optimal Transport with an frame-pair-conditioned dustbin score $s$, predicted by a dustbin score predictor. Finally, the global density map of each frame is decomposed into shared density map $\hat{\mathbf{D}}^s_t$ and $\hat{\mathbf{D}}^s_{t+\delta}$ (not visualized) and outflow/inflow density maps $\hat{\mathbf{D}}^o_t$ and $\hat{\mathbf{D}}^{in}_{t+\delta} $ through group-wise association.}
	\caption{The overall pipeline of GD\textsuperscript{3}A and GIA-Track. During descriptor extraction and enhancement, multiple descriptors are extracted for each pedestrian head. Using a learnable dustbin query, the Dustbin Score Predictor derives a frame-pair-conditioned dustbin score $s$ from the pedestrian descriptors in the current frame pair, enabling Optimal Transport to better distinguish cross-frame identities and establish more accurate pixel-level descriptor matches. These pixel-level descriptor matches are subsequently used to decompose the global density maps into shared, inflow, and outflow components through group-wise density assignment and to establish instance-level identity associations through group-wise identity association for dense crowd tracking.}
	\label{fig:pipeline}
\end{figure*}

\subsection{Problem Formulation and Overall Framework}
Given a video clip $V=\{F_i\}_{i=1}^{n}$ captured by a moving drone in a scene with dense crowds, the goal is to count the number of unique pedestrians $M(V)$ appearing throughout the clip and track each pedestrian. For counting, we estimate the global density map $\hat{\mathbf{D}}^g_1$ of the first frame $F_1$ and the inflow density maps $\hat{\mathbf{D}}^{in}_i$ for all subsequent frames.  The global density map $\hat{\mathbf{D}}^g_t$  contains the density values of all pedestrians in frame $F_t$, whereas the inflow density map $\hat{\mathbf{D}}^{in}_t$ contains the density values of pedestrians that newly appear in $F_{t}$ compared to $F_{t-1}$. $M(V)$ can then be computed using the following formulation:
\begin{equation}
	M(V) \approx \text{sum}(\hat{\mathbf{D}}^g_1) + \sum_{k=1}^{(n/\delta)-1} \text{sum}(\hat{\mathbf{D}}^{in}_{1+k\times\delta}),
\end{equation}
where $\delta$ denotes the sampling interval between frames. For tracking, it needs to determine the position coordinates and identity of each pedestrian in every frame.

The overall pipeline is illustrated in Fig. \ref{fig:pipeline}. The training set $\mathcal{V}=\{V_j, P_j, ID_j\}_{j=1}^m$ consists of $m$ video clips $V_j$ along with their corresponding annotations. The annotations include the coordinates of pedestrians' head $P_j$ in each frame and their unique identity $ID_j$ throughout the entire video clip. During training, frames $F_t$ and $F_{t+\delta}$ are sampled from a video clip $V_j$ with a random interval $\delta$. The feature maps $\mathbf{F}_t$ and $\mathbf{F}_{t+\delta}$, as well as the global density maps $\hat{\mathbf{D}}_t^g$ and $\hat{\mathbf{D}}_{t+\delta}^g$, are then obtained using the backbone and a pre-trained image-level density estimation model, respectively. The feature maps are first filtered using the global density maps to retain visual descriptors $\{\mathbf{f}^t_i\}_{i=1}^N$ and $\{\mathbf{f}^{t+\delta}_i\}_{i=1}^M$ for each pedestrian head in each frame. Note that a group of multiple visual descriptors, rather than a single descriptor, is retained for each pedestrian head. These visual descriptors are then enhanced using position. The enhanced descriptors are processed by an Attentional Graph Neural Network (AGNN)~\citep{SuperGlue} to obtain association descriptors $\{\mathbf{d}^t_i\}_{i=1}^N$ and $\{\mathbf{d}^{t+\delta}_i\}_{i=1}^M$. Association descriptors from the two frames, together with the dustbin query, are fed into the dustbin score predictor to obtain a frame-pair-conditioned dustbin score $s$. Optimal transport then incorporates the frame-pair-conditioned dustbin score $s$ to solve optimal matching matrix $\mathbf{P}^*$ between association descriptors. 

Finally, based on $\mathbf{P}^*$, GD\textsuperscript{3}A performs GDA to decompose the global density map $\hat{\mathbf{D}}_t^g$ into the shared density map $\hat{\mathbf{D}}_t^s$ and the outflow density map $\hat{\mathbf{D}}_t^{o}$. Similarly, $\hat{\mathbf{D}}_{t+\delta}^g$ is decomposed into the shared density map $\hat{\mathbf{D}}_{t+\delta}^s$ and the inflow density map $\hat{\mathbf{D}}_{t+\delta}^{in}$. The shared density map $\hat{\mathbf{D}}_{t}^s$ ($\hat{\mathbf{D}}_{t+\delta}^s$) represents pedestrians that appear in both frames $F_t$ and $F_{t+\delta}$. 
The outflow density map $\hat{\mathbf{D}}_t^{o}$ contains pedestrians present in $F_t$ but absent in $F_{t+\delta}$.
Note that the shared and outflow density maps are byproducts, and only the inflow density map is useful. For tracking, 
pedestrian coordinates are obtained by detecting local maxima in the global density map. Based on the pixel-level descriptor matches, GIA-Track employs GIA to convert them into instance-level pedestrian associations. Next, we provide a detailed description of each component.
\subsection{Pixel-Level Descriptor Matching}
\subsubsection{Descriptor Extraction and Enhancement}
For two given consecutive frames $F_t$ and $F_{t+\delta}$, their feature maps are extracted by the backbone:
\begin{equation}
	\resizebox{0.8\hsize}{!}{$\mathbf{F}_t = \mathrm{backbone}(F_t), \quad \mathbf{F}_{t+\delta} = \mathrm{backbone}(F_{t+\delta}).$}
	\label{eq:feature_extraction}
\end{equation}
The dimensions of the feature maps $\mathbf{F}_t$ and $\mathbf{F}_{t+\delta}$ are $\mathbb{R}^{\frac{H}{r} \times \frac{W}{r}} $, where $H$ and $W$ are the height and width of the input frame, and $r$ is the downsampling rate. Each feature map contains $\frac{H}{r} \times \frac{W}{r}$ visual descriptors. Associating all descriptors across two frames incurs prohibitive computational costs, and most descriptors correspond to the background regions, making their involvement unnecessary. Thus, we filter the feature maps using predicted global density maps obtained through a pre-trained image-level counter:
\begin{equation}
	\hat{\mathbf{D}}^g_t = \mathrm{counter}(F_t), \quad \hat{\mathbf{D}}^g_{t+\delta} = \mathrm{counter}(F_{t+\delta}).
	\label{eq:global_density_map}
\end{equation}
The filter process can be described as:
\begin{equation}
	{\mathbf{F}}' = \mathbf{F} \odot \mathbb{I}(\hat{\mathbf{D}}^g > \tau),
	\label{eq:filter}
\end{equation}
where $\mathbb{I}$ is the indicator function and $\tau$ is the pre-defined threshold. Filtered feature maps $\mathbf{F}_t'$ and $\mathbf{F}_{t+\delta}'$ only contain visual descriptors of pedestrian heads, and the association of these descriptors significantly reduces the computational cost.  

Formally, we define the set $\mathcal{A}_t = \{\mathbf{f}_i^t, \mathbf{p}_i^t\}_i^N$, where each $\mathbf{f}_i^t \in \mathbb{R}^D$ is a non-zero descriptor in $\mathbf{F}_t'$, and $\mathbf{p}_i^t = (x_i^t,y_i^t)$ is the coordinate of $\mathbf{f}_i^t$ in $\mathbf{F}_t'$. Similarly, $\mathcal{B}_{t+\delta} = \{\mathbf{f}_i^{t+\delta}, \mathbf{p}_i^{t+\delta}\}_i^M$ denotes the corresponding set constructed from ${\mathbf{F}}'_{t+\delta}$.

Due to the high similarity of pedestrian head appearances, accurately associating descriptors belonging to the same pedestrian across two frames is challenging. Therefore, visual descriptors are first enriched with spatial positions, as the same pedestrian typically appears at nearby locations in adjacent frames. To this end, an encoder is utilized to project the 2D vector $(x_i^t,y_i^t)$ into the same feature space as the visual descriptor $\mathbf{f}$, and then the element-wise addition is performed between the projected vector and the visual descriptor:
\begin{equation}
	^{(0)}\mathbf{f}_i^t = \mathbf{f}_i^t \oplus  \mathrm{Encoder}(x_i^t,  y_i^t).
	\label{eq:add_position}
\end{equation}

To further enhance the distinctiveness of the descriptors, we use an AGNN to aggregate spatial and visual contextual cues via an iterative message-passing mechanism, which alternates between intra-image self-attention to encode local relationships and inter-image cross-attention to resolve matching ambiguities.
% The AGNN consisting of $L$ layers, formed by alternating self-attention and cross-attention layers, is utilized to aggregate contextual cues for each descriptor. 
The computation of the $i$-th descriptor at the $l$-th layer is:
\begin{equation}
	^{(l+1)}\mathbf{f}_i^t =  {}^{(l)}\mathbf{f}_i^t + \mathrm{MLP}([^{(l)}\mathbf{f}_i^t || ^{(l)}\tilde{\mathbf{f}}]),
	\label{eq:add_contex_cues}
\end{equation}
where $^{(l)}\tilde{\mathbf{f}}$ is the message aggregated from other descriptors of the current frame or adjacent frame:
\begin{equation}
	\setlength\abovedisplayskip{4pt}
	\setlength\belowdisplayskip{4pt}
	\resizebox{0.8\hsize}{!}{$
		\begin{aligned}
			\mathbf{Q}={}^{(l)}\mathbf{f}_i^t&\mathbf{W}^{Q}, \hspace{0.2cm} \mathbf{K}=\tilde{\mathbf{F}}^l_\theta \mathbf{W}^{K}, \hspace{0.2cm} \mathbf{V}=\tilde{\mathbf{F}}^l_\theta\mathbf{W}^{V}, \\
			^{(l)}\tilde{\mathbf{f}} &= \mathrm{Softmax}(\frac{\mathbf{Q}\mathbf{K}^T}{\sqrt{D}})\mathbf{V},\\
		\end{aligned}
		$}
	\label{eq:attention}
\end{equation}
where $\mathbf{W}$ represents the learnable parameters at each layer. Note that each layer has its own learnable parameters, but the superscript $l$ on $\mathbf{W}$ is omitted for brevity. $\tilde{\mathbf{F}}^l_\theta$ is the matrix obtained by concatenating all descriptors from the $l$-th layer of the current frame or adjacent frame.
When $l$ is even, $\theta$ is set to $t$, and $\tilde{\mathbf{F}}^l_t = [^{(l)}\mathbf{f}_0^t;{}^{(l)}\mathbf{f}_1^t;...;{}^{(l)}\mathbf{f}_{N-1}^t]$. When $l$ is odd, $\theta$ is set to $t+\delta$, and $\tilde{\mathbf{F}}^l_{t+\delta} = [{}^{(l)}\mathbf{f}_0^{t+\delta};{}^{(l)}\mathbf{f}_1^{t+\delta};...;{}^{(l)}\mathbf{f}_{M-1}^{t+\delta}]$. Thus, Eq. \ref{eq:attention} alternately apply self-attention and cross-attention to aggregate information from descriptors in the current frame or the neighboring frame.
After passing through $L$ layers, the output features $^{(L)}\mathbf{f}^t_i$ and $^{(L)}\mathbf{f}_i^{t+\delta}$ are obtained, which are then fed into a linear layer to produce the final descriptors for association:
\begin{equation}
	\setlength\abovedisplayskip{3pt}
	\setlength\belowdisplayskip{3pt}
	\resizebox{0.7\hsize}{!}{$
		\mathbf{d}_i^t=\mathrm{MLP}(^{(L)}\mathbf{f}^t_i), \quad \mathbf{d}_i^{t+\delta}=\mathrm{MLP}(^{(L)}\mathbf{f}^{t+\delta}_i).
		$}
	\label{eq:mathcing_descriptor}
\end{equation}
% By applying Equations (4), (5), and (6) to all visual features in $\mathcal{A}$ and $\mathcal{B}$, we obtain descriptors $^{(L)}\mathbf{f}_t$ and $^{(L)}\mathbf{f}_{t+\delta}$ for association.

\subsubsection{Pixel-level Descriptor Matching via OT with a Frame-Pair-Conditioned Dustbin Score}
Inspired by feature matching and graph matching~\citep{SuperGlue}, we perform augmented optimal transport on pixel-level descriptors for matching. An additional dustbin is introduced to match descriptors of inflow and outflow pedestrians that only appear in one frame. The dustbin score acts as a threshold to distinguish whether a descriptor corresponds to a pedestrian appearing in both frames. Previous methods typically learn a dustbin score for an entire dataset. However, such a strategy fails to account for the features of pedestrians in the current input, making it suboptimal. In contrast, we design a dustbin score predictor that infers a frame-pair-conditioned dustbin score from the pedestrian descriptors of each input frame pair. Descriptor association can be formulated as \textbf{maximizing}  the following objective:
\begin{equation}
	\label{eq:OT}
	\setlength\abovedisplayskip{4pt}
	\setlength\belowdisplayskip{4pt}
	\resizebox{0.8\hsize}{!}{$
		\begin{aligned}
			\mathrm{L}(\mathbf{a}, \mathbf{b}) &= \max_{P \in \mathbf{U}(\mathbf{a}, \mathbf{b})}\langle\mathbf{C}, \mathbf{P}\rangle \\
			&= \max_{P \in \mathbf{U}(\mathbf{a}, \mathbf{b})} \sum_{i \in \llbracket N+1\rrbracket,j\in \llbracket M+1 \rrbracket} \mathbf{C}_{i,j}\mathbf{P}_{i,j}
		\end{aligned}
		$}
\end{equation}
where $\mathbf{C}$ is the cost matrix and is composed as:
\begin{equation}
	\mathbf{C}=
	\begin{bmatrix}
		\mathbf{S}_{N\times M} & \mathbf{s}_{N \times 1} \\
		\mathbf{s}_{1 \times M} & s
	\end{bmatrix},
	\label{eq:cost_matrix}
\end{equation}
where $\mathbf{S}_{ij} = \langle\mathbf{d}_i^t,\mathbf{d}_j^{t+\delta}\rangle $ is the similarity of the descriptors. The entries of $\mathbf{s}_{N \times 1}$ and $\mathbf{s}_{1 \times M}$ are all set to the frame-pair-conditioned dustbin score $s$ that is computed as:
\begin{equation}
	\begin{gathered}
		\mathbf{X}_{in} = [\mathbf{q}, \mathbf{d}_1^t, \dots, \mathbf{d}_N^t, \mathbf{q}, \mathbf{d}_1^{t+\delta}, \dots, \mathbf{d}_M^{t+\delta}], \\
		\mathbf{X}_{out} = \operatorname{TransformerEncoder}(\mathbf{X}_{in}), \\
		\mathbf{s}^1, \mathbf{s}^2 = \mathbf{X}_{out}[1], \mathbf{X}_{out}[N+2], \\
		s = \operatorname{MLP}(\operatorname{Concat}(\mathbf{s}^1, \mathbf{s}^2)),
	\end{gathered}
	\label{eq:adaptive_score}
\end{equation}
where $\mathbf{q}$ is the learnable dustbin query. 

In Eq. \ref{eq:OT}, $\mathbf{P}$ is the matching matrix to be solved, and the feasible set of $\mathbf{P}$ is defined as follows:
\begin{equation}
    \mathbf{U}(\mathbf{a},\mathbf{b})=\{\mathbf{P}\mid\mathbf{P}\mathbbold{1}=\mathbf{a},\ \mathbf{P}^{\top}\mathbbold{1}=\mathbf{b}\},
\end{equation}
where $\mathbf{a}$ and $\mathbf{b}$ denote the marginal distributions, which are set as $\left[\mathbbold{1}_{N}^{\top}, M\right]^{\top}$ and $\left[\mathbbold{1}_{M}^{\top}, N\right]^{\top}$ for matching.
For $i\leq N$ and $j \leq M$, $\mathbf{P}_{ij}$ is the probability of matching the $i$-th descriptor from frame $t$ with the $j$-th descriptor from frame $t+\delta$. When $i=N+1$ and $j \leq M$, $\mathbf{P}_{ij}$ represents the probability that the $j$-th descriptor in frame $t+\delta$ matches the dustbin (i.e., belongs to an inflow pedestrian), while for $j=M+1$ and $i \leq N$, $\mathbf{P}_{ij}$ signifies the probability that the $i$-th descriptor in frame t matches the dustbin (i.e., belongs to an outflow pedestrian).
Eq. \ref{eq:OT} is essentially a linear programming problem with $N+M+2$ equality constraints, and the optimal matrix $\mathbf{P}^*$ can be obtained by Sinkhorn iteration~\citep{Sinkhorn}.
\begin{figure}
	\centering
	\includegraphics[width=0.45\textwidth]{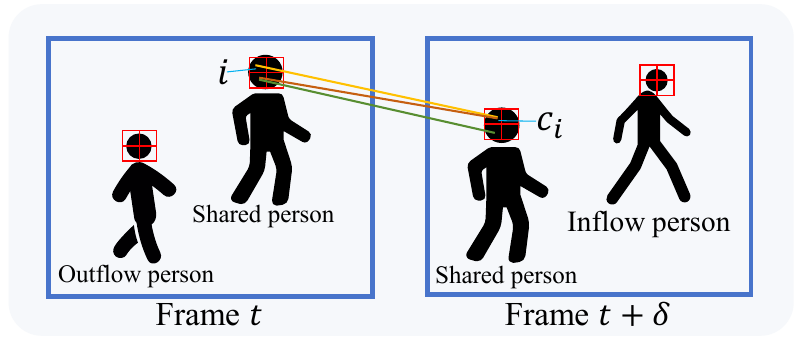}
	\setlength{\abovecaptionskip}{-2pt}% local override for this tightly cropped figure
	\caption{Illustration of group-wise density assignment. For the \(i\)-th descriptor \(\mathbf{f}_i^t\) in frame \(t\), its highest matching score points to the \(c_i\)-th descriptor \(\mathbf{f}_{c_i}^{t+\delta}\) in frame \(t+\delta\). However, the best-matched descriptor of \(\mathbf{f}_{c_i}^{t+\delta}\) in frame \(t\) may not be \(\mathbf{f}_i^t\). We regard this match as valid if \(i\) is among the Top-\(K\) descriptors with the highest matching scores to \(\mathbf{f}_{c_i}^{t+\delta}\), since intra-group spatial mismatches do not necessarily alter the corresponding density assignment. By contrast, descriptors of inflow and outflow pedestrians do not satisfy the reverse Top-\(K\) criterion. Their densities are therefore assigned to the inflow/outflow density maps.}
	\label{fig:group-wise_association}
\end{figure}
\subsection{GD\textsuperscript{3}A: Global Density Map Decomposition via Group-wise Density Assignment}
By solving Eq. \ref{eq:OT}, the optimal matching matrix $\mathbf{P}^*$ is obtained. Based on $\mathbf{P}^*$, the predicted global density map can be decomposed into inflow, outflow, and shared density maps.
Since each pedestrian head contains a group of multiple descriptors, it is reasonable that a descriptor at the center of a head in the current frame matches a descriptor located at the top-right of the same pedestrian's head in the adjacent frame. In other words, we allow matching errors within the descriptor group of each pedestrian, which improves robustness without compromising the accuracy of the final results. Based on this observation, we adopt a reverse top-$K$ association strategy to implement GDA. The group size \(K\) is determined by the kernel size of the image-level density estimator and the downsampling ratio \(r\), i.e., \(K=(\text{kernel size}/r)^2\). Specifically, for the $i$-th ($ 1\le i \leq N $) descriptor $\mathbf{f}_i^t$ in frame $F_t$ (the $i$-th row in $\mathbf{P}^*$), the column index of its best-matched descriptor in $F_{t+\delta}$ is:
\begin{equation}
	c_i = \arg\max_{c\in\{1,\ldots,M\}}\mathbf{P}^*_{i,c},
	\label{eq:max_column}
\end{equation}
and the set of row indices corresponding to the top-$K$ largest values in column $c_i$ is defined as:
\begin{equation}
	\mathcal{R}^{c_i}_{\mathrm{top}K}
	= \left\{r\in\{1,\ldots,N\}\mid
	\mathbf{P}^*_{r,c_i}\geq v_{K}^{c_i}\right\},
	\label{eq:topk}
\end{equation}
where $v_{K}^{c_i}$ is the $K$-th largest value in column $c_i$. An intuitive illustration of this process is provided in Figure~\ref{fig:group-wise_association}.
Based on Eq. \ref{eq:max_column} and \ref{eq:topk}, the index $c^*_i$ of descriptor in $F_{t+\delta}$ matched to $\mathbf{f}_i^t$ can be obtained:
\begin{equation}
	c^*_i = \begin{cases} 
		c_i, & \begin{array}{l}
			\text{if } i \in \mathcal{R}^{c_i}_{\mathrm{top}K},
		\end{array}\\
		-1, &  \text{otherwise}
	\end{cases}
	\label{eq:find_matched_descriptor_index}
\end{equation}
If $ c^*_i \neq -1$, the density value corresponding to $\mathbf{f}_i^t$ in the global density map $\hat{\mathbf{D}}^g_t$ is assigned to the shared density map $\hat{\mathbf{D}}_t^s$; otherwise, it is assigned to the outflow density map $\hat{\mathbf{D}}_t^o$. Similarly, global density map $\hat{\mathbf{D}}^g_{t+\delta}$ can be decomposed into shared density map $\hat{\mathbf{D}}_{t+\delta}^s$ and inflow density maps $\hat{\mathbf{D}}_{t+\delta}^{in}$. 

\subsection{GIA-Track: Group-wise Identity Association Tracker}
By detecting local maxima in the global density map $ \hat{\mathbf{D}}^g_t$ and $ \hat{\mathbf{D}}^g_{t+\delta}$ and extracting their corresponding coordinates $\{\tilde{\mathbf{p}}^t_k\}_{k=1}^{N_p} $ and $\{\tilde{\mathbf{p}}^{t+\delta}_k\}_{k=1}^{M_p} $, the positions of all pedestrians can be obtained, where $N_p$ and $M_p$ denote the numbers of pedestrians in frames $t$ and $t+\delta$, respectively. For a descriptor in $F_t$ or $F_{t+\delta}$, its corresponding pedestrian can be identified using the following formulation:
\begin{equation}
	k_i = \arg\min_{k}d(\mathbf{p}_i,\tilde{\mathbf{p}}_k),
	\label{eq:find_pedestrian_index}
\end{equation}

% Declare the compact float here so it enters the queue while page 12's left
% column is being built and is placed at the top of the right column.
\begin{algorithm}[!t]
	\caption{\footnotesize Pseudocode of GD\textsuperscript{3}A and GIA-Track.}
	\label{alg:Pseudocode}
	\footnotesize
	\begin{algorithmic}[1]
		\Require Video frames $F_t$ and $F_{t+\delta}$
		\Ensure Inflow density map $\hat{\mathbf{D}}_{t+\delta}^{in}$, pedestrian trajectories

		\State \textbf{Stage 1: Descriptor Extraction and Enhancement}
		\State Extract features $\mathbf{F}_t, \mathbf{F}_{t+\delta}$ (Eq. \ref{eq:feature_extraction}) and global density maps $\hat{\mathbf{D}}^g_t, \hat{\mathbf{D}}^g_{t+\delta}$ (Eq. \ref{eq:global_density_map}).
		\State Filter features to obtain descriptor sets:
		\Statex \quad $\mathcal{A}_t = \{\mathbf{f}_i^t, \mathbf{p}_i^t\}_{i=1}^N$ and $\mathcal{B}_{t+\delta} = \{\mathbf{f}_j^{t+\delta}, \mathbf{p}_j^{t+\delta}\}_{j=1}^M$.
		\State Enhance descriptors to get $\mathbf{d}_i^t$ and $\mathbf{d}_j^{t+\delta}$ (Eq. \ref{eq:add_position}--\ref{eq:mathcing_descriptor}).

		\State \textbf{Stage 2: Pixel-Level Descriptor Matching}
		\State Compute frame-pair-conditioned dustbin score $s$ (Eq. \ref{eq:adaptive_score})
		\State Construct the cost matrix $\mathbf{C}$ (Eq. \ref{eq:cost_matrix})
		\State Solve OT and obtain optimal matching matrix $\mathbf{P}^*$ (Eq. \ref{eq:OT})

		\State \textbf{Stage 3: Task-Specific Group-wise Aggregation}
		\State Get pedestrian positions $\{\tilde{\mathbf{p}}^t_k\}^{N_p}_{k=1}$ and $\{\tilde{\mathbf{p}}^{t+\delta}_k\}_{k=1}^{M_p}$.

		\State Initialize identity-association matrix $\mathbf{V}=\mathbf{0}_{N_p\times M_p}$.
		\For{each descriptor $i$ in both frame directions}
		\State Get match index $c^*_i$ (Eq. \ref{eq:max_column}--\ref{eq:find_matched_descriptor_index}).
		\If{$c^*_i \neq -1$ (Matched)}
		\State \textit{GDA:} assign density to the \textit{Shared} map.

		\State \textit{GIA:} obtain $k_i$ and $k_{c^*_i}$ (Eq. \ref{eq:find_pedestrian_index}).
		\State Update $\mathbf{V}_{k_i, k_{c^*_i}} \leftarrow \mathbf{V}_{k_i, k_{c^*_i}} + 1$.
		\Else
		\State \textit{GDA:} assign density to an inflow/outflow map.
		\EndIf
		\EndFor

		\State Apply Hungarian matching to $\mathbf{V}$ and propagate IDs.
		\State Assign new IDs to unmatched pedestrians.

		\State \Return Inflow density map $\hat{\mathbf{D}}_{t+\delta}^{in}$, updated trajectories
	\end{algorithmic}
\end{algorithm}

where $d$ denotes the distance between two points. Let the identity-association score matrix be $\mathbf{V} \in \mathbb{R}^{N_p \times M_p}$, for each descriptor $\mathbf{f}_i^t$ in frame $F_t$, we first obtain the index $c^*_i$ of its matched descriptors in frame $F_{t+\delta}$ using Eq. \ref{eq:find_matched_descriptor_index}. If $ c^*_i \neq -1$,  their corresponding pedestrian indices $k_i$ and $k_{c^*_i}$ are then determined using Eq. \ref{eq:find_pedestrian_index}, and the corresponding entry $\mathbf{V}_{k_i,k_{c^*_i}}$ in the identity-association score matrix is incremented by one. By applying the same procedure to all descriptors $\mathbf{f}_j^{t+\delta}$ in $F_{t+\delta}$, the final identity-association score matrix $\mathbf{V}$ is obtained. The Hungarian algorithm~\citep{Hungarian} is then applied to 
$\mathbf{V}$ to derive the pedestrian associations between the two frames. Based on these associations, pedestrian IDs from $F_{t}$ are propagated to $F_{t+\delta}$, while new IDs are assigned to pedestrians in $F_{t+\delta}$ that remain unmatched (i.e., those with corresponding entries equal to 0 in the identity-association score matrix $\mathbf{V}$). Algorithm~\ref{alg:Pseudocode} summarizes the execution process of GD\textsuperscript{3}A and GIA-Track. Please refer to it for more details.

\subsection{Loss Function}
Since the dataset provides only the coordinates of pedestrian head centers and their corresponding identity labels, pixel-level descriptor correspondences are unavailable. Therefore, we first extend the point-level annotations to pixel-level annotations for the head regions based on local spatial correspondence. Assume that, for a given pedestrian, the head center is located at $\mathbf{p}_t$ in frame $t$ and at $\mathbf{p}_{t+\delta}$ in frame $t+\delta$. Using local spatial correspondence, we can infer pixel-level correspondence for the surrounding local region based on a local displacement offset $\mathbf{\Delta} \in \mathbb{Z}^2$:
\begin{equation}
	\mathbf{p}_t+\mathbf{\Delta} \longleftrightarrow \mathbf{p}_{t+\delta}+\mathbf{\Delta}, \forall \mathbf{\Delta} \in \mathbb{Z}^2,\|\mathbf{\Delta}\|_\infty < \rho,
\end{equation}
where $\rho$ is the pre-defined radius of the local region, and $\longleftrightarrow$ indicates that the descriptors at the two positions correspond to each other (i.e., the same local position of the same pedestrian head in two frames).
Using the above extension, the indices of descriptors to be matched between the two frames are divided into three sets: $\mathcal{M}$, $\mathcal{U}_A$, and $\mathcal{U}_B$. $\mathcal{M}=\{(i,j)_k\}_{k=1}^{N_m}$ contains the indices of matched descriptors, $\mathcal{U}_A$ contains the indices of descriptors belonging to outflow pedestrians, and $\mathcal{U}_B$ contains the indices of descriptors belonging to inflow pedestrians. Finally, the loss can be computed as follows:
\begin{equation}
	\begin{aligned}
		\mathcal{L} = &-\sum_{(i,j)\in \mathcal{M}} \log \mathbf{P}_{i,j} \\
		&- \sum_{i\in \mathcal{U}_A} \log \mathbf{P}_{i,M+1} - \sum_{j\in \mathcal{U}_B} \log \mathbf{P}_{N+1,j},
	\end{aligned}
\end{equation}
where $\mathbf{P}$ is the matrix obtained by solving Eq.~\ref{eq:OT}.

Thanks to the differentiability of the optimal transport algorithm, backpropagation can be performed from the association stage to the backbone and the dustbin score predictor, thereby encouraging descriptors of the same pedestrian to be similar, those of different pedestrians to be dissimilar, and enabling the dustbin score predictor to learn a frame-pair-conditioned dustbin score from the descriptors of each input frame pair.

\section{Experiments}

\begin{table*}[!t]
	\centering
	\caption{Comparative results for video-level crowd counting on MovingDroneCrowd++. Clips are divided into four difficulty levels $D_0$ $\sim$ $D_3$ based on the number of unique pedestrian, with the trajectories ranges of [0, 200), [200, 400), [400, 600), and $\geq$ 600, respectively. Our method achieves the best overall performance, with clear advantages on high-difficulty clips.}
	\resizebox{\textwidth}{!}{%
		\begin{tabular}{l|c|ccc|cc|cccc}
			\toprule
			\multirow{2}{*}{Method} & \multirow{2}{*}{Venue} & \multirow{2}{*}{MAE$\downarrow$} & \multirow{2}{*}{RMSE$\downarrow$} & \multirow{2}{*}{WRAE$\downarrow$} & \multirow{2}{*}{MIAE$\downarrow$} & \multirow{2}{*}{MOAE$\downarrow$} & \multicolumn{4}{c}{Density levels} \\
			\cmidrule(lr){8-11}
			& & & & & & &  \raisebox{-0.5ex} {$D_0$} & \raisebox{-0.5ex} {$D_1$} & \raisebox{-0.5ex} {$D_2$} & \raisebox{-0.5ex} {$D_3$} \\
			\midrule
			\multicolumn{11}{l}{\textit{\textbf{Multi-Object Tracking Methods}}} \\
			ByteTrack\citep{ByteTrack} & ECCV'22 & 244.32 & 551.05 & 117.39 & 18.35 & 18.17 & 92.07 & 227.67 & 364.20 & 1448.00 \\
			BoT-SORT\citep{BoT_SORT} & arXiv'22 & 278.02 & 589.75 & 132.99 & 20.92 & 20.80 & 131.41  & 215.83 & 368.80 & 1570.67 \\
			OC-SORT\citep{OC-SORT} & CVPR'23 & 188.49 & 308.74 & 75.85 & 10.43 & 11.27 & 61.89 & 268.17 &  423.00 & 777.67 \\
			DiffMOT\citep{DiffMOT} & CVPR'24 & 337.85  & 802.59 & 165.91  & 25.87 & 25.58 & 129.52  & 248.83 & 571.80 & 2001.00 \\
			MOTIP\citep{MOTIP} & CVPR'25 & 116.61 & 215.57  &  47.72 & 8.87 & 8.03 & 51.37 & 103.33 & 163.20 & 652.67  \\
			\midrule
			\multicolumn{11}{l}{\textit{\textbf{Localization-based VIC Methods}}} \\
			DRNet\citep{DRNet} & CVPR'22 & 83.04 & 172.07 & 30.88 & 8.60 & 7.96 & 28.23 & 96.41 & 160.67 & 420.20 \\
			CGNet\citep{uavvic} & CVPR'24 & 80.85  & 184.51 & \underline{26.13} & -- & -- & \underline{19.37} & 96.17  & 171.40  & 452.67  \\
			\midrule
			LOI\citep{LOI} & ECCV'16 & 245.24 & 357.76 & 99.08  & -- & -- & 103.20 & 328.72 & 476.90  & 970.48 \\
			\midrule
			\multicolumn{11}{l}{\textit{\textbf{Density map-based VIC Methods}}} \\
			FMDC\citep{FMDC} & WACV'24 & 127.78 & 208.82 & 46.22& 7.69& 7.44& 46.10& 190.62 & 235.93& 556.93 \\
			SDNet\citep{MDC} (Ours) & ICCV'25 & \underline{76.24}  & \underline{160.33} & 32.40  & \underline{6.40} & \underline{6.08} & 31.87 & \underline{88.84} & \underline{143.68} & \underline{337.96}  \\ \midrule
			% GD\textsuperscript{3}A(VGG16) & -- & \checkmark & 60.08 & 91.72 & 30.20 & 4.35 & 3.82 & 35.41 & 96.93 & 47.42 & 229.54 \\
			% GD\textsuperscript{3}A(ResNet50) & -- & \checkmark & 57.47 & 91.95 &  28.51 & 4.26 & 4.36 & 32.70 & 75.74 & 59.32 & 240.82 \\
			\multirow{2}{*}{GD\textsuperscript{3}A(Ours)} & \multirow{2}{*}{--} & \textbf{40.11 }& \textbf{71.61}& \textbf{18.83} & \textbf{3.67}& \textbf{3.47} & \textbf{17.96}& \textbf{66.30} & \textbf{83.94} & \textbf{114.10} \\
			& & \textcolor{red}{$\downarrow$ 47.4\%}  & \textcolor{red}{$\downarrow$ 55.3\%} & \textcolor{red}{$\downarrow$ 27.9\%}   & \textcolor{red}{$\downarrow$ 42.7\%} & \textcolor{red}{$\downarrow$ 42.9\%} & \textcolor{red}{$\downarrow$ 7.3\%} & \textcolor{red}{$\downarrow$ 25.4\%}  & \textcolor{red}{$\downarrow$ 41.6\%}  & \textcolor{red}{$\downarrow$ 66.2\%} \\
			\bottomrule
		\end{tabular}
	}
	\label{tab:comparison_counting_on_mdc++}
\end{table*}

This section first introduces the datasets and evaluation metrics, followed by the key implementation details. We then compare our method with various related approaches to demonstrate its superior performance. Ablation studies further verify the robustness of our method, and visualization comparisons intuitively highlight its interpretability compared to existing methods.
\subsection{Experiment Setup}
\subsubsection{Datasets}
To validate the effectiveness of the proposed method, we conduct experiments on both our MovingDroneCrowd++ captured by moving drones and the surveillance video dataset VSCrowd~\citep{VSCrowd}. VSCrowd contains 634 video clips recorded by fixed surveillance cameras across 153 different scenes, with a resolution of $1920 \times 1080$ and a total of 62,938 frames. We adopt the same dataset split for training and evaluation in DRNet~\citep{DRNet}. MovingDroneCrowd++ and VSCrowd provide sufficiently diverse data to enable a comprehensive evaluation of different approaches.
\subsubsection{Evaluation Metric}
The primary evaluation metrics for counting are video-level MAE and RMSE:
\begin{equation}
	\resizebox{0.84\columnwidth}{!}{$\displaystyle
		\mathrm{MAE}=\frac{1}{N}\sum_{i=1}^{N}\left|y_i-\hat{y}_i\right|,\quad
		\mathrm{RMSE}=\sqrt{\frac{1}{N}\sum_{i=1}^{N}\left(y_i-\hat{y}_i\right)^2}.
	$}
\end{equation}
where $y_i$ denotes the ground-truth count of distinct pedestrians in a video clip, $\hat{y}_i$ is the predicted count, and $N$ is the number of video clips in test set.
In addition, we adopt WRAE, MIAE, and MOAE defined in DRNet~\citep{DRNet}. WRAE (Weighted Relative Absolute Errors) weights the relative error by the proportion of frames in each video clip, thereby accounting for the impact of video lengths.
% \begin{equation}
	% \text{WRAE} = 
	% \sum_{i=1}^{N} 
	% \frac{T_i}{\sum_{j=1}^{N} T_j}
	% \cdot
	% \frac{\lvert y_i - \hat{y}_i \rvert}{y_i}
	% \times 100\%,
	% \end{equation}
% where $T_i$ is the number of frames in the $i$-th clip. 
The previously mentioned metrics evaluate errors at the video level, whereas MIAE and MOAE measure the errors of pedestrian inflow and outflow at the frame pair level.
% \begin{equation}
	% \scriptsize
	% \resizebox{0.92\columnwidth}{!}{$
		% \text{MIAE} =
		% \frac{\displaystyle \sum_{i=1}^{N}\sum_{t=\delta}^{T_i-1}
			% \left|\hat{y}_i^{+}(t) - y_i^{+}(t)\right|}
		% { \sum_{i=1}^{N}(T_i - \delta)}, \quad
		% \text{MOAE} =
		% \frac{\displaystyle \sum_{i=1}^{N}\sum_{t=\delta}^{T_i-1}
			% \left|\hat{y}_i^{-}(t) - y_i^{-}(t)\right|}
		% { \sum_{i=1}^{N}(T_i - \delta)},
		% $}
	% \end{equation}
% where $\hat{y}_i^+(t)$ and $y_i^+(t)$ are the predicted and ground-truth inflow counts of frame $t$ compared to frame $t-\delta$, respectively. Similarly, $\hat{y}_i^-(t)$ and $y_i^-(t)$ denote the predicted and ground-truth pedestrian outflow counts from frame $t-\delta$ to frame $t$, respectively. These two metrics compute the average absolute errors of pedestrian inflow and outflow for each pair of adjacent frames. 
For tracking, we adopt the widely used metric HOTA~\citep{HOTA}, which provides a balanced evaluation of detection (DetA) and association (AssA) accuracy. In addition, we also report the results of MOTA~\citep{MOTA} and IDF1~\citep{IDF1}.

\subsubsection{Implementation Details}
During training, the frame interval is randomly sampled to ensure diverse pedestrian motion patterns and varying drone speeds. In data augmentation, random horizontal flipping is not applied, as it would disrupt the positional consistency of the same pedestrian across frames. The cropping and scaling strategies follow DRNet~\citep{DRNet}. We adopt ResNet50~\citep{ResNet50} as the backbone network (initialized with weights pretrained on ImageNet), followed by an FPN~\citep{FPN} to enhance multi-scale representation. 
%  During training, features are filtered using ground-truth global density maps, whereas during test, filtering relies on density maps predicted by a pre-trained counter. This design allows the model to focus on learning pedestrian descriptor matching during training stage. 
The initial learning rate is set to 5e-5 for the backbone and 1e-4 for the dustbin score predictor and AGNN. The model is implemented in PyTorch and trained on RTX 3090 with a global batch size of 8.
\subsection{Comparison with State-of-the-Art Methods}
In this subsection, we compare the proposed GD\textsuperscript{3}A and GIA-Track with several representative SOTA methods on video individual counting and multi-object tracking tasks.
\begin{table}[!t]
	\centering
	\caption{Comparison of multi-object tracking on MovingDroneCrowd++. Higher values are better. GIA-Track achieves the best performance and significantly outperforms previous methods.}
	\scriptsize
	\setlength{\tabcolsep}{1pt}
	\begin{tabular}{l|ccccc}
			\toprule
			Method & HOTA & DetA & AssA & MOTA & IDF1 \\ \midrule
			ByteTrack & 17.5 & 20.8 & 17.1 & \underline{5.0}  & 11.7 \\
			BoT-SORT & 17.5 & 20.8& 16.9 & 4.8 & 11.4 \\
			OC-SORT &  11.7 & 6.3 & \underline{24.7} & 3.0 & 7.3 \\
			DiffMOT & \underline{20.9} & \underline{28.2} & 17.6 & 4.6 & \underline{14.2} \\
			MOTIP & 11.2 & 14.3 & 10.0 & -7.7 & 8.5 \\ \midrule
			\multirow{2}{*}{GIA-Track} & \textbf{34.4} & \textbf{29.4} & \textbf{41.2} & \textbf{12.8} & \textbf{41.3} \\
			&\textcolor{red}{+64.6\%}&\textcolor{red}{+4.3\%}&\textcolor{red}{+66.8\%}&\textcolor{red}{+156\%}&\textcolor{red}{+190.8\%} \\
			\bottomrule
	\end{tabular}
	\label{tab:comparison_tracking_on_MDC++}
\end{table}
\subsubsection{Comparison of Video Individual Counting on MovingDroneCrowd++} 
As shown in Table \ref{tab:comparison_counting_on_mdc++}, different types of methods are separated by horizontal lines. Multi-object tracking (MOT) methods count pedestrians by tracking each individual in the video and using the number of resulting trajectories as the final count. However, MOT methods perform poorly on MovingDroneCrowd++, especially on highly challenging video clips. 
The best MOT method, MOTIP, still produces much larger errors than GD\textsuperscript{3}A in terms of MAE/RMSE, i.e., 116.61/215.57 vs. 40.11/71.61. 
On the most difficult density level \(D_3\), the gap further increases, with MOTIP obtaining 652.67 MAE compared with 114.10 MAE of our method. 
These results indicate that MOT-based methods struggle to maintain reliable trajectories under dense crowd scenes and rapid drone motion. Localization-based VIC methods significantly improve over MOT-based methods, but they still suffer from limited counting accuracy on MovingDroneCrowd++. 
CGNet reports an MAE/RMSE of 80.85/184.51, while DRNet obtains 83.04/172.07. 
Moreover, both methods show large errors on the most challenging density level \(D_3\), with 420.20 for DRNet and 452.67 for CGNet. 
This suggests that strict localization and one-to-one association errors remain severe in dense crowds with complex illumination and small pedestrian heads. Density map-based VIC methods avoid explicit pedestrian localization and association by directly estimating density maps. 
FMDC directly predicts inflow and outflow density maps between consecutive frames, but this challenging paradigm leads to limited performance, with 127.78 MAE and 208.82 RMSE, even falls behind localization-based approaches.
Our conference method, SDNet, alleviates the task complexity by first estimating shared density maps, improving the results to 76.24 MAE and 160.33 RMSE.
However, accurately estimating the density map between two frames is challenging. As time progresses, erroneous density estimates gradually accumulate, thereby degrading the final counting performance.

In contrast, our method GD\textsuperscript{3}A avoids the strict localization and one-to-one association process and achieves SOTA performance by decoupling global density maps into shared, outflow and inflow components via robust GDA. Notably, the performance gains become more pronounced as the video difficulty increases: compared with previous methods, GD\textsuperscript{3}A reduces the counting error by 41.6\% and 66.2\% on the high-difficulty subsets $D_2$ and $D_3$.
\subsubsection{Comparison of Video Individual Crowd Counting on VSCrowd} 
In addition to experiments on our dynamic drone video dataset, we also compare our method with other approaches on large-scale surveillance video dataset VSCrowd. As illustrated in Table \ref{tab:comparison_on_VSCrowd}, our method achieves the best overall performance on datasets captured by fixed surveillance cameras. This indicates that our approach is effective not only in moving drone scenarios but also in static surveillance settings, highlighting its remarkable generalizability and robustness. On this dataset, the performance gap between our approach and existing methods is less pronounced, primarily because localization and one-to-one association is relatively easier from a fixed surveillance perspective. The reduced scene complexity allows localization-based methods to achieve competitive results with ours.
\subsubsection{Comparison of Tracking on MovingDroneCrowd++}
% Table \ref{tab:comparison_tracking_on_MDC++} presents a comparison between our method, GIA-Track, and recent classical and state-of-the-art multi-object tracking methods on MovingDroneCrowd++. These methods include both end-to-end Transformer-based method and conventional tracking-by-detection methods. Experimental results demonstrate that our method significantly outperforms all competing methods in dense crowd scenarios captured by moving drones. Existing methods suffer from either poor detection performance or weak association capability under dense crowds with complex motion conditions. In particular, the Transformer-based MOT method MOTIP performs poorly because it relies on a predefined vocabulary size for identity representation, making it unsuitable for dense crowd scenarios in our dataset, where scaling the vocabulary size leads to prohibitive training costs. Overall, our method GIA-Track achieves a 64.6\% improvement in HOTA over the second-best method DiffMOT, highlighting its strong superiority in dense and complex motion scenarios.
Table \ref{tab:comparison_tracking_on_MDC++} compares GIA-Track with recent state-of-the-art multi-object tracking methods, including end-to-end Transformer-based and conventional tracking-by-detection approaches. Existing methods suffer from poor detection or weak association under dense crowds and complex motion. In particular, MOTIP relies on a predefined identity vocabulary, whose expansion to dense scenes incurs prohibitive training costs. By contrast, GIA-Track aggregates descriptor-level matching evidence before assigning pedestrian identities, thereby reducing the risk that isolated descriptor mismatches directly cause incorrect pedestrian associations or affect the identity assignments of other pedestrians in dense scenes. It consequently improves HOTA by 64.6\% over the second-best method DiffMOT, demonstrating greater robustness in dense crowds under complex drone motion.

\begin{table*}[htbp]
	\caption{Comparative video-level crowd counting results on surveillance video dataset VSCrowd demonstrate that our method achieves the best overall performance. This indicates that our approach performs well on both moving drone and fixed surveillance scenarios. D0 $\sim$ D4  denote five pedestrian density range: [0, 50), [50, 100), [100, 150), [150, 200), $\geq 200$, respectively.}
	\centering
	\resizebox{\textwidth}{!}{%
	\begin{tabular}{l|c|ccc|cc|ccccc}
		\toprule
		\multirow{2}{*}{Method} & \multirow{2}{*}{Venue} & \multirow{2}{*}{MAE$\downarrow$} & \multirow{2}{*}{RMSE$\downarrow$} & \multirow{2}{*}{WRAE(\%)$\downarrow$} & \multirow{2}{*}{MIAE$\downarrow$} & \multirow{2}{*}{MOAE$\downarrow$} & \multicolumn{5}{c}{Density levels} \\
		\cmidrule(lr){8-12}
		& & & & & & & \raisebox{-0.5ex} {$D_0$} & \raisebox{-0.5ex} {$D_1$} & \raisebox{-0.5ex} {$D_2$} & \raisebox{-0.5ex} {$D_3$} & \raisebox{-0.5ex} {$D_4$} \\
		\midrule
		FairMOT \citep{Zhang2021FairMOT} & IJCV'21  & 35.4 & 62.3 & 48.9 & 4.9 & 4.4 & 13.5 & 22.4 & 67.9 & 84.4 & 145.8  \\
		HeadHunter-T \citep{Headhunter-T} & CVPR'21 & 30.0 & 50.6 & 38.6 & 4.0 & 4.1 & 11.8 & 25.7 & 56.0 & 92.6 & 131.4  \\
		\midrule
		LOI \citep{LOI} & ECCV'16 & 24.7 & 33.1 & 37.4 & -- & -- & 12.5 & 25.4 & 39.3 & 39.6 & 86.7  \\
		\midrule
		DRNet \citep{DRNet} & CVPR'22 & 9.3 & 16.5 & 12.1 & \underline{2.0} & \underline{2.0} & 4.9 & 7.3 & 13.9 & 30.6 & 42.3  \\
		CGNet \citep{uavvic} & CVPR'24 & 8.9 & 17.7 & 12.6 & -- & -- & 5.0 & \underline{5.8} & \textbf{8.5} & 25.0 & 63.4 \\
		PDTR \citep{PGDTR} & MM'24 & 9.6 & 17.6 & 11.4 & -- & --& 4.6 & 6.8 & 14.7 & 23.6 & 60.6 \\
		OMAN \citep{OMAN} & ICIP'25 & \underline{8.3}  & \underline{15.4} & 11.1  &-- &-- & \underline{3.9} &6.2  & 13.2 &\underline{20.3}  &\underline{48.2}  \\ \midrule
		SDNet \citep{MDC} (Ours) & ICCV'25 & 8.6  & 15.9 & \underline{11.0}  & --& --& 4.0 &6.6  & 12.6 &24.9  &\textbf{47.9}  \\
		
		GD\textsuperscript{3}A(Ours) & -- & \textbf{7.5} & \textbf{15.4} & \textbf{8.9} & \textbf{1.7} & \textbf{1.8} & \textbf{3.3}   & \textbf{5.1}  & \underline{11.2} & \textbf{17.3} & 53.8 \\
		\bottomrule
	\end{tabular}
	}
	\label{tab:comparison_on_VSCrowd}
\end{table*}
\subsection{Ablation Studies}
\begin{table}[!t]
	\centering
	\caption{Ablation study on the effects of pedestrian position (Pos.), density (Den.), and AGNN during matching. Lower metric values are better.
		% This indicates that enhancing visual descriptors with pedestrian location and aggregating contextual information via AGNN during matching can improve the discriminability of matching descriptors, thereby boosting performance. In contrast, discrepancies in pedestrian density values between training and testing introduce noise, which degrades the final performance.
	}
	\label{tab:ablation_association}
	\scriptsize
	\setlength{\tabcolsep}{2pt}
	\begin{tabular}{cccc|ccc}
			\toprule
			Set. & Pos. & Den. & AGNN & MAE & RMSE & WRAE \\ \midrule
			S0 & & & & 71.37 & 120.03 & 34.14 \\
			S1 &\ding{51} & & & 69.62 & 115.36& 32.72 \\
			S2 & & \ding{51} & & 72.29 & 124.38& 35.42 \\
			S3 & \ding{51} & \ding{51} & &  70.01 & 117.48 & 33.68  \\
			S4 & & & \ding{51} & 68.59 &  108.66 & 34.21 \\
			\textbf{S5} &\ding{51} &         & \ding{51}     
			& \textbf{66.13} & \textbf{101.70} & \textbf{30.50}    \\ 
			S6 &    & \ding{51}        & \ding{51}     
			& 69.90& 123.88& 31.88    \\ 
			S7 & \ding{51} & \ding{51} & \ding{51}
			& 66.61 &  111.01 &  30.71      \\ \bottomrule
	\end{tabular}
\end{table}

\subsubsection{Effect of Position and Density on the Association}
We first conduct ablation studies on the pixel-level descriptor matching process by examining the effects of pedestrian locations, density values, and contextual aggregation with AGNN. As shown in Table \ref{tab:ablation_association}, pedestrian locations and AGNN provide clear benefits for descriptor matching. Compared with the baseline setting S0, combining pedestrian locations and AGNN in S5 reduces MAE/RMSE from 71.37/120.03 to 66.13/101.70, achieving the best performance. In contrast, density values are less effective: using density alone in S2 even performs worse than the baseline, and further adding density to S5, i.e., S7, increases MAE/RMSE from 66.13/101.70 to 66.61/111.01. These results indicate that pedestrian locations provide discriminative spatial priors, while AGNN enhances descriptor discriminability through contextual aggregation. 
By contrast, density values mainly reflect local crowd distribution rather than pedestrian identity, and thus may introduce ambiguity in dense scenes.

\begin{table}[!t]
	\caption{Ablation study of the frame-pair-conditioned dustbin score on the validation set of MovingDroneCrowd++, where FPCDS denotes the frame-pair-conditioned dustbin score.}
	\centering
	\begin{tabular}{ccccc}
		\toprule
		\multicolumn{1}{l}{Method} & FPCDS & MAE $\downarrow$ & RMSE $\downarrow$ & WRAE $\downarrow$  \\ \midrule
		\multirow{2}{*}{DRNet}     & \ding{55} & 91.79     & 131.47      & 40.09      \\
		& \checkmark & \textbf{77.71}    &  \textbf{120.65}    & \textbf{27.91}   \\ \midrule
		\multirow{2}{*}{Ours}      & \ding{55}  & 54.14     & 102.23      & 23.82       \\
		& \checkmark  & \textbf{32.25}   & \textbf{46.58}     & \textbf{18.84}       \\ \bottomrule
	\end{tabular}
	\label{tab:adaptive_dustbin_score_ablation}
\end{table}

\subsubsection{Effect of the Frame-Pair-Conditioned Dustbin Score}
Table \ref{tab:adaptive_dustbin_score_ablation} presents an ablation study on the effect of the frame-pair-conditioned dustbin score. We conduct experiments on the OT-based instance-level matching method DRNet and our pixel-level descriptor matching method GD\textsuperscript{3}A, comparing a dataset-level learnable dustbin score with the proposed frame-pair-conditioned dustbin score. The results show that using the frame-pair-conditioned dustbin score significantly improves the final counting performance. This indicates that the frame-pair-conditioned dustbin score can effectively distinguish shared pedestrians from inflow and outflow ones based on the pedestrian descriptors in the current frame pair.

% Declare the ablation figure after the first two ablation analyses so that it
% is placed at the top of page 17, before the qualitative visualizations.
\begin{figure*}[!t]
	\centering
	\includegraphics[width=\linewidth]{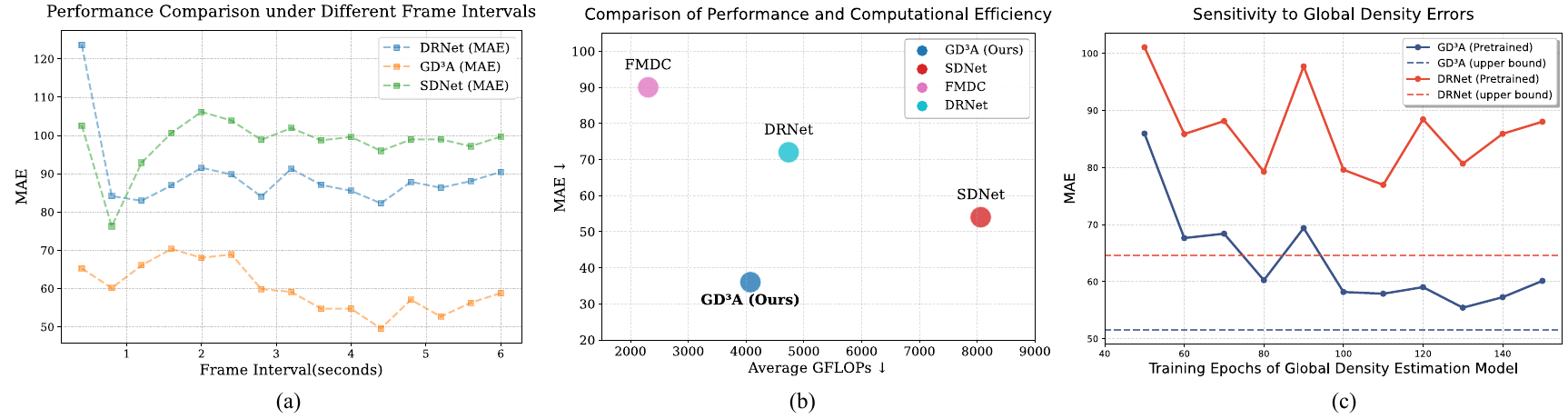}
	\caption{(a) Comparison of our method with other methods under different frame intervals, ranging from \SIrange{0.04}{6}{\second} with a step size of 0.04 s. (b) Performance--efficiency trade-off comparison between our method and existing approaches. (c) Sensitivity analysis to errors in the predicted density maps for our method and other approaches.}
	\label{fig:ablation_visual}
\end{figure*}

\begin{table}[!t]
	\caption{Effect of Group-wise Density Assignment (GDA) and Group-wise Identity Association (GIA). Lower counting errors and higher tracking scores are better.}
	\label{tab:ablation_Reverse_Top_k_Descriptor_Voting}
	\centering
	\scriptsize
	\setlength{\tabcolsep}{1.5pt}
	\begin{tabular}{lccccc}
		\toprule
		\multicolumn{6}{c}{\textbf{Group-wise Density Assignment}} \\
		\midrule
		Method & MAE & RMSE & WRAE & MIAE & MOAE \\
		\midrule
		GD\textsuperscript{3}A
		& \textbf{40.11} & \textbf{71.61} & \textbf{18.83}
		& \textbf{3.67} & \textbf{3.47} \\
		w/o GDA
		& 129.98 & 236.87 & 58.43 & 10.10 & 9.56 \\
		\midrule
		\multicolumn{6}{c}{\textbf{Group-wise Identity Association}} \\
		\midrule
		Method & HOTA & DetA & AssA & MOTA & IDF1 \\
		\midrule
		GIA-Track
		& \textbf{34.4} & \textbf{29.4} & \textbf{41.2}
		& \textbf{12.8} & \textbf{41.3} \\
		w/o GIA
		& 22.7 & 29.4 & 19.3 & -7.2 & 20.2 \\
		\bottomrule
	\end{tabular}
\end{table}

\subsubsection{Effect of Group-wise Density Assignment and Group-wise Identity Association}
% Table~\ref{tab:ablation_Reverse_Top_k_Descriptor_Voting} shows the effects of GDA in GD\textsuperscript{3}A and GIA in GIA-Track. In GD\textsuperscript{3}A w/o GDA, we adopt a strict one-to-one association strategy similar to DRNet~\citep{DRNet}, which leads to a significant performance drop. This indicates that GDA is more suitable for dense crowds and complex illumination conditions, as its intra-group error tolerance effectively mitigates error accumulation. GIA-Track w/o GIA directly matches the descriptors at peak points of pedestrian heads and transfers IDs across adjacent frames according to the matching results, resulting in a large drop in ID association metrics (AssA). This demonstrates that GIA is more robust for ID association than directly relying on single-descriptor matching.
Table~\ref{tab:ablation_Reverse_Top_k_Descriptor_Voting} shows the effects of GDA in GD\textsuperscript{3}A and GIA in GIA-Track. In GD\textsuperscript{3}A w/o GDA, we adopt a strict one-to-one association strategy similar to DRNet~\citep{DRNet}, which leads to a substantial degradation in counting performance. This result supports the benefit of aggregating pixel-level descriptors matching evidence through GDA, which reduces sensitivity to intra-group spatial mismatches and their accumulation as counting errors. GIA-Track w/o GIA directly matches the descriptors at pedestrian-head peak points and transfers identities across adjacent frames according to these single-descriptor matches, resulting in a large drop in AssA. In contrast, GIA aggregates multiple descriptor correspondences for each pedestrian before determining pedestrian-level identities, thereby reducing the impact of isolated cross-identity descriptor mismatches and limiting their propagation to other pedestrian assignments.

% Queue the two qualitative figures after the ablation figure. Their combined
% height keeps them on consecutive pages 18 and 19.
\begin{figure*}[!t]
	\centering
	\includegraphics[width=\linewidth]{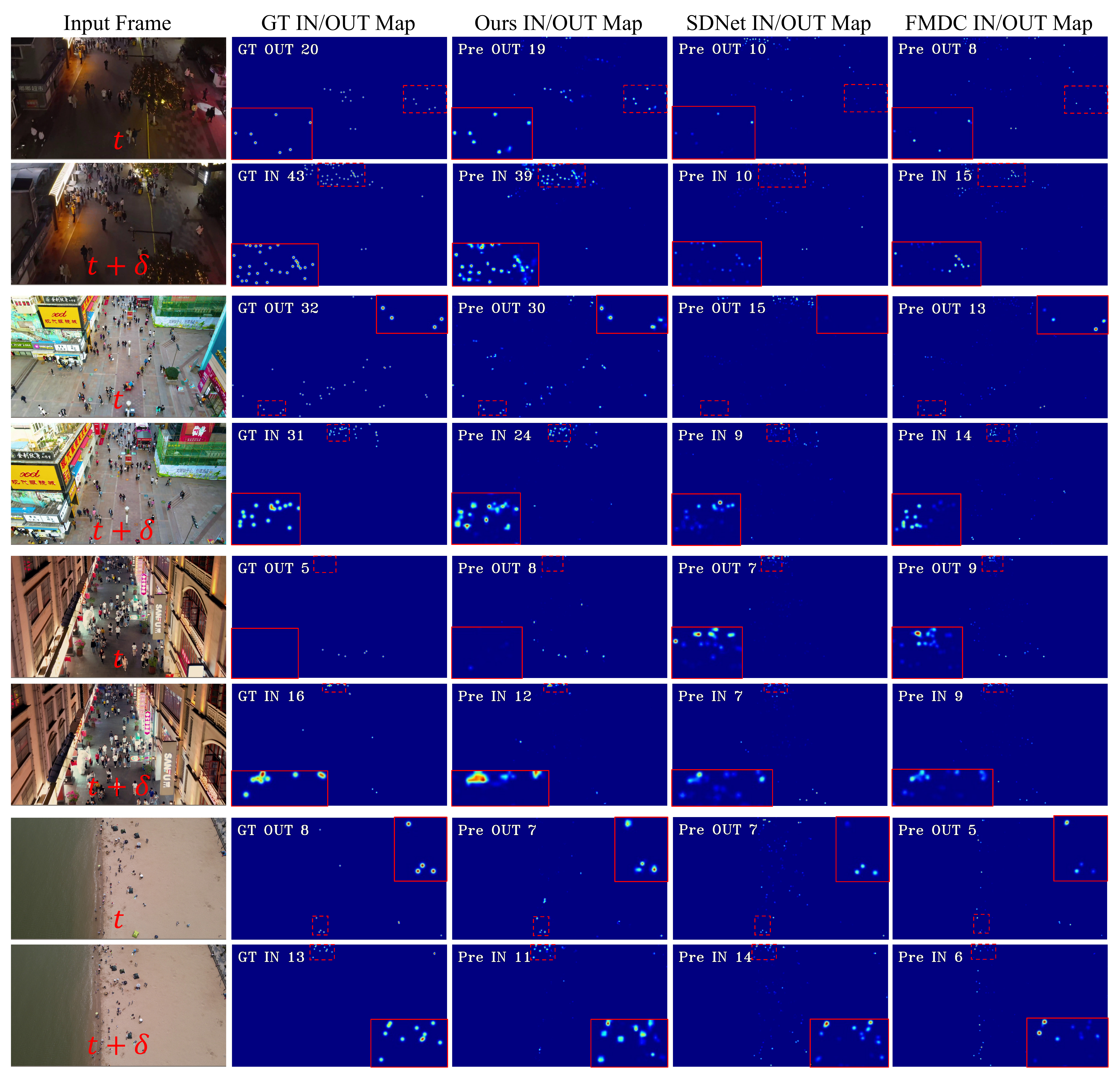}
	\caption{Visual comparison of inflow and outflow density maps predicted by our method and other density map-based VIC methods. For each frame pair, the first row shows the outflow density maps, while the second row presents the inflow density maps. Compared with other methods, our approach more accurately predicts inflow and outflow counts and yields density maps that are more interpretable and more consistent with the ground-truth.}
	\label{fig:visual}
\end{figure*}
\begin{figure*}[!t]
	\centering
	\includegraphics[width=\linewidth]{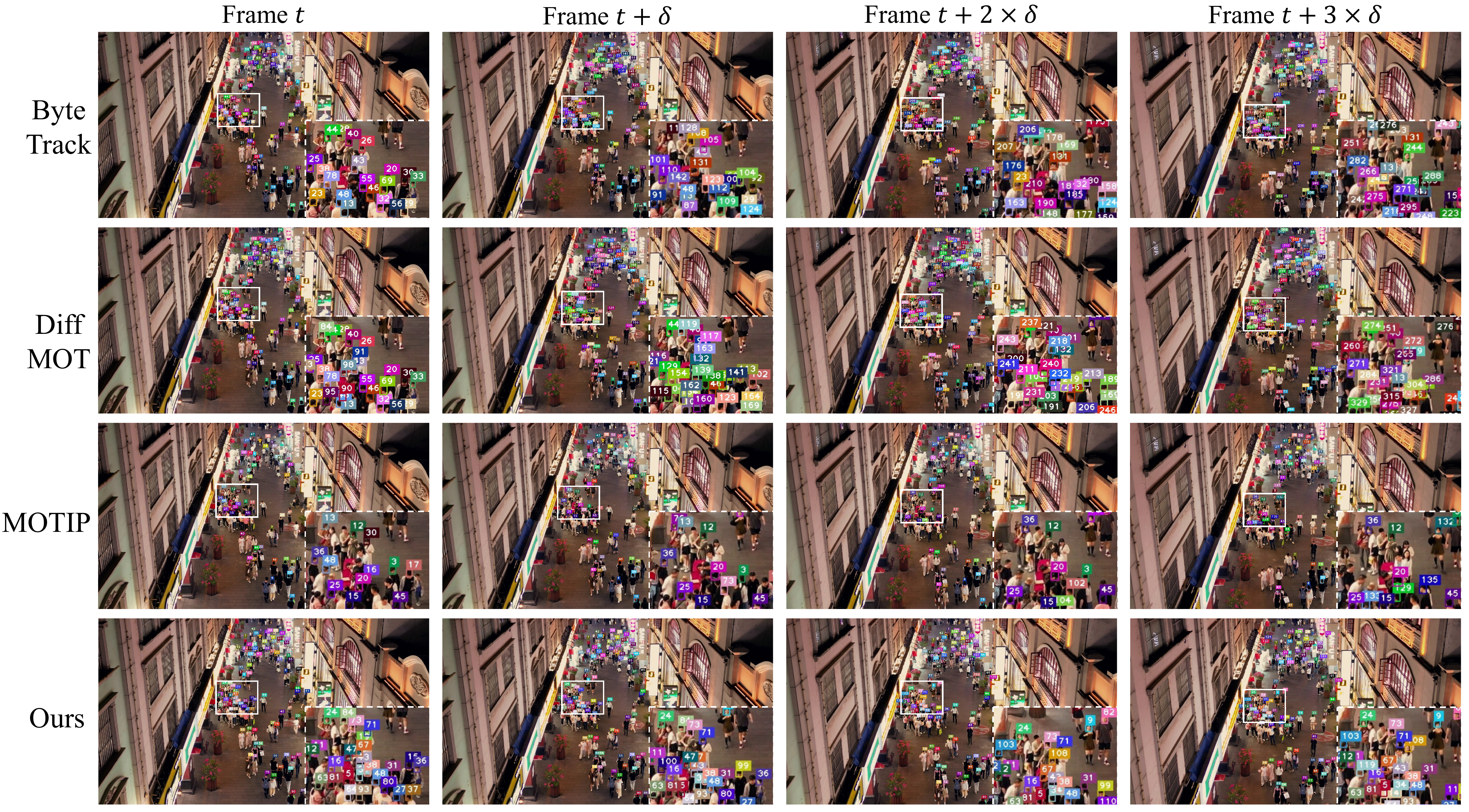}
	\caption{Visual comparisons of tracking results. Other methods suffer from frequent ID switches and localization errors, whereas our method GIA-Track maintains more consistent identities. The enlarged white dashed boxes show the local details of the regions indicated by the solid white boxes.}
	\label{fig:track_visual}
\end{figure*}

\subsubsection{Effect of Frame Sampling Interval at Test Time}
To evaluate the sensitivity of our method to temporal intervals and its performance under different drone movement speeds, we test our method and competing methods under frame sampling intervals ranging \SIrange{0.04}{6}{\second} with a step size of 0.04s. For our dataset captured by moving drones, this range covers a wide spectrum of temporal variations and thus provides a comprehensive evaluation. As shown in Fig. \ref{fig:ablation_visual}(a), our method exhibits stable performance across the entire interval range. In contrast, the other methods achieve their best performance at an interval of approximately 1 second, after which their performance degrades noticeably as the interval increases. This indicates that our method is robust to temporal variations at test time. Even with substantial variations in the frame interval, the performance remains stable without significant fluctuations. More importantly, it remains reliable when drones move at high speeds.

\subsubsection{Trade-off between Performance and Efficiency}
Fig. \ref{fig:ablation_visual}(b) illustrates the performance--efficiency trade-off. Our method achieves the best balance between performance and efficiency, delivering the highest accuracy while maintaining high computational efficiency. Compared with our conference method SDNet, which is based on cross-attention, GD\textsuperscript{3}A filters feature maps using global density maps to exclude descriptors from dominant background regions. This significantly reduces redundant computations and improves computational efficiency.

\subsubsection{Sensitivity to Global Density Estimation}
Fig. \ref{fig:ablation_visual}(c) presents a sensitivity analysis of our method and the localization-based method DRNet with respect to the predicted global density maps. The red dashed line indicates the counting MAE of DRNet when ground-truth global density maps are provided at test time, which allows DRNet to use accurate pedestrian locations and thus eliminates localization errors (error lower bound). The red solid lines denote the MAE of DRNet when using global density maps predicted by models trained for different numbers of epochs. 
As shown, the red solid line exhibits large fluctuations and a clear gap from the upper bound, indicating that localization-based methods are highly sensitive to the accuracy of density and localization predictions. In contrast, our method GD\textsuperscript{3}A (represented by the blue curves) is considerably more robust to density estimation errors, showing only a small performance gap between using predicted density maps and the upper-bound performance obtained using ground-truth density maps.

\subsection{Qualitative Results}
To intuitively demonstrate the superiority of our method over other methods, Fig. \ref{fig:visual} presents a visual comparison of the inflow and outflow density maps predicted by our method and other density map-based VIC methods on two adjacent frames. For each frame pair, the first row shows the outflow density maps, while the second row presents the inflow density maps. The ground-truth or predicted inflow/outflow counts are presented in the top-left corner of each density map. Red dashed boxes indicate representative regions, while the red solid boxes show enlarged views of the corresponding regions. As can be observed, compared with other methods, our approach achieves superior performance in terms of both the interpretability of the predicted inflow and outflow density maps and the accuracy of the corresponding inflow and outflow counts. This advantage stems from the fact that our method can more effectively distinguish between inflow and outflow pedestrians, whereas other methods struggle to do so, resulting in noisier predicted inflow and outflow density maps.

% Fig. \ref{fig:track_visual} presents qualitative comparisons between GIA-Track and other SOTA multi-object tracking methods. As shown in the highlighted regions within the enlarged white dashed boxes, tracking-by-detection methods can detect more pedestrians but suffer from severe identity switches across frames due to the rapid motion of the drone. In contrast, transformer-based methods exhibit fewer identity switches but miss a large number of pedestrians. Our method detects most pedestrians while maintaining high identity consistency across frames. These results indicate that existing MOT methods degrade significantly under complex motion and dense crowd conditions, whereas our approach handles such scenarios more effectively.
Fig. \ref{fig:track_visual} presents qualitative comparisons between GIA-Track and other SOTA multi-object tracking methods. As shown in the enlarged white dashed boxes, tracking-by-detection methods detect more pedestrians but suffer from severe identity switches under rapid drone motion, whereas transformer-based methods exhibit fewer identity switches but miss many pedestrians. Our method detects most pedestrians with fewer identity switches, consistent with its aggregation of descriptor-level matching evidence, which limits isolated cross-identity descriptor errors from propagating into trajectory-level identity errors under complex motion and dense crowds.

\section{Conclusion}
This paper presents MovingDroneCrowd++, a challenging benchmark for VIC and MOT in large-scale scenes with dense crowds captured by moving drones, together with two accurate and robust methods GD\textsuperscript{3}A and GIA-Track for the two tasks. Both methods are unified by cross-frame pixel-level descriptor correspondences established via OT with a frame-pair-conditioned dustbin score. Based on these correspondences, GD\textsuperscript{3}A decomposes global density maps into shared, inflow, and outflow density maps through group-wise density assignment, whereas GIA-Track establishes pedestrian identities and trajectories through group-wise identity association. By aggregating multiple descriptor correspondences for each pedestrian, both methods tolerate intra-group spatial mismatches and limit the propagation of isolated inter-group descriptor errors. Experiments demonstrate substantial gains under dense crowds and complex camera motion, reducing the counting error by 47.4\% and improving tracking performance by 64.6\%. These improvements validate  the effectiveness of aggregating multiple pixel-level descriptor matches for each pedestrian, which enables both methods to effectively handle  such challenging conditions. Overall, our dataset and methods further bridge the gap between theoretical research and practical applications.
\section*{Statements and Declarations}

\bmhead{Funding}
This work was supported in part by the JC STEM Lab of AI for Science and Engineering, funded by The Hong Kong Jockey Club Charities Trust; in part by the MTR Research Funding (MRF) Scheme under Grant CHU-24003; and in part by the Research Grants Council of Hong Kong under Project CUHK14213224.

\bmhead{\mbox{Author Contributions}}
Yaowu Fan contributed to all stages of the work, from conceiving the initial idea to methodology development, algorithm implementation, experimentation, result analysis, and preparation of the final draft. Jia Wan contributed primarily to proposing and conceptualizing the original idea and planning the experiments. Tao Han contributed to discussions of the algorithmic implementation details and provided substantial constructive suggestions. Andy J. Ma, Wanli Ouyang, and Antoni B. Chan contributed primarily to revising the manuscript and discussing and interpreting the experimental results.

\bmhead{\mbox{Competing Interests}}
The authors have no competing interests to declare that are relevant to the content of this article.

\bmhead{\mbox{Data and Code Availability}}
The MovingDroneCrowd++ dataset and its annotations, together with the source code and pretrained models for GD\textsuperscript{3}A and GIA-Track, are publicly available through the project repository at \url{https://github.com/fyw1999/MovingDroneCrowd}.

\bibliography{main}% common bib file

\end{document}